\documentclass[10pt,twocolumn,letterpaper]{article}

\usepackage{cvpr}              

\usepackage{url}

\usepackage{xspace}
\usepackage{dsfont}
\usepackage{afterpage}

\usepackage{enumitem}
\usepackage{booktabs}
\usepackage{caption} 
\usepackage{multirow} 
\usepackage{enumitem}
\usepackage{colortbl} 
\usepackage{bbm}
\usepackage{algorithm}
\usepackage{algpseudocode}                  
\usepackage{setspace}                       
\usepackage{lipsum} 
\usepackage{tikz}

\usepackage{mathtools} 
\usepackage{bm} 
\usepackage{soul} 
\usepackage{nicefrac}
\usepackage{csquotes}
\usepackage{array}
\newcolumntype{L}[1]{>{\raggedright\let\newline\\\arraybackslash\hspace{0pt}}m{#1}}
\newcolumntype{C}[1]{>{\centering\let\newline\\\arraybackslash\hspace{0pt}}m{#1}}
\newcolumntype{R}[1]{>{\raggedleft\let\newline\\\arraybackslash\hspace{0pt}}m{#1}}
\newcolumntype{s}{>{\columncolor{gray!15}}c}

\usepackage{duckuments}
\usepackage{wrapfig}

\newcommand\blfootnote[1]{
  \begingroup
  \renewcommand\thefootnote{}\footnote{#1}
  \addtocounter{footnote}{-1}
  \endgroup
}

\definecolor{cvprblue}{rgb}{0.21,0.49,0.74}
\usepackage[pagebackref,breaklinks,colorlinks,allcolors=cvprblue]{hyperref}

\title{SSAFE: Simple and Strong AI-Generated Image Detection \\
via Frozen Vision Encoders}

\author{
    Seunghyun Lee$^{1, \dagger}$ \hspace{6pt}
    Byoungkwon Kim$^{1}$ \hspace{6pt}
    Jaehyun Nam $^{2, \dagger}$ \hspace{6pt}
    Kyungmin Lee $^{1}$ \hspace{6pt} 
    Jinwoo Shin$^{1}$  \\[0.3em]
    $^{1}$KAIST \hspace{10pt}
    $^{2}$Google Cloud AI
    \\
    \tt\small shyun4839@kaist.ac.kr, shyun4839@gmail.com
}

\begin{document}
\maketitle
\begin{abstract}

The rapid advancement of generative models has blurred the boundary between synthetic and real imagery, creating an urgent need for reliable deepfake detection. Yet most existing approaches rely on massive real--fake datasets, which are increasingly difficult to maintain as new generators continue to emerge. In this work, we investigate how much information about image authenticity is already encoded in modern multimodal vision representations. We find that frozen multimodal encoders naturally separate real and synthetic images in their embedding space, enabling a simple linear classifier to achieve strong performance without task-specific fine-tuning. Motivated by this observation, we develop a representation-aware data curation strategy that selects a compact set of representative generators for training. The resulting training set contains only 10K images, compared to 288K in AIGIBench and 4M in OpenFake, while improving robustness to unseen generators and distribution shifts. We additionally introduce RealWorldBench, a benchmark consisting of modern camera photographs, contemporary stock images, and outputs from recent commercial generators. Experiments across multiple benchmarks show that combining frozen multimodal representations with carefully curated training data provides a simple and effective approach to AI-generated image detection.

\blfootnote{
$^{\dagger}$ Work done at KAIST.
}


\end{abstract}

    
\vspace{-16pt}
\section{Introduction}
\label{sec:intro}

The rapid advancement of text-to-image (T2I) generative models, such as Imagen~\citep{deepmind2024imagen3}, Flux~\citep{blackforestlabs2024flux1dev}, Nano-Banana~\citep{google2025gemini25flashimage}, has enabled the synthesis of highly realistic images that are nearly indistinguishable from real photographs~\citep{livernoche2025openfake, lu2023seeing}. While these models opened new application possibilities in fields like creative content generation~\citep{openai2024dalle3, deepmind2024imagen3, blackforestlabs2024flux1dev, wang2024instantid, li2024photomaker, midjourney2024v61}, they simultaneously raise significant concerns regarding visual information~\citep{lu2023seeing} and deepfakes~\citep{wang2020cnn, rossler2019faceforensics++}. Consequently, this has necessitated the development of reliable systems for detecting AI-generated images~\citep{wang2020cnn, yan2024sanity, li2024improving, yan2024orthogonal, liang2025ferretnet, wang2025lota}.

\begin{figure}
    \centering\small
    \includegraphics[width=0.98\linewidth]{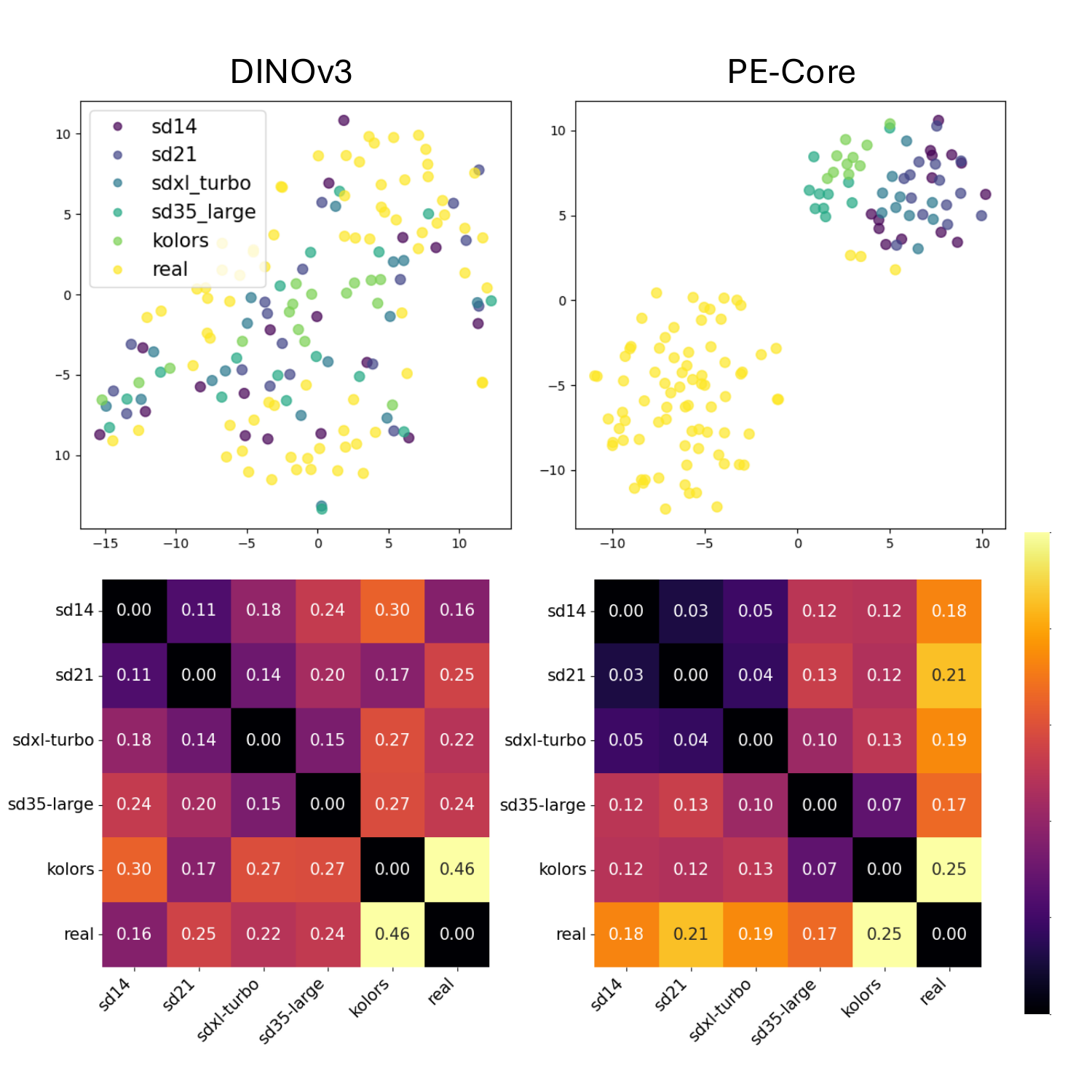}
    \vspace{-0.5cm}
    \caption{t-SNE visualization of embeddings from two vision encoders.
DINOv3-ViTL16 (left) shows overlapping clusters between real and synthetic images, while PE-Core-G14-448 (right) clearly separates real images and reveals higher-level grouping structures among generators. This highlights the stronger discriminative and generator-aware representation of pretrained encoders.}
    \label{fig:tsne_3}
    \vspace{-0.5cm}
\end{figure}


A common approach to building such systems involves fine-tuning pre-trained Vision-Language Models (VLMs) using both AI-generated and real photographs~\citep{zhou2025aigi, tan2025c2p, ojha2023towards}. However, we observe that these approaches often fail to generalize to synthetic images generated by unseen T2I models during fine-tuning. For example, recent state-of-the-art detectors trained on Stable-Diffusion-v1-4 (SD-v1.4~\citep{rombach2022high}) and ProGAN~\citep{karras2018progressive} achieve only slightly better accuracy than random guessing on the in AIGIBench~\citep{li2025artificial}. Furthermore, continuously collecting training data from all existing T2I models is computationally burdensome and requires constant updates as new T2I models emerge, making this approach impractical.

To alleviate such issues, we first revisit AI-generated image detection from the representation perspective of multimodal models. Specifically, we perform an in-depth analysis of the embedding spaces of CLIP-series~\citep{radford2021learning}, SigLIP-series~\citep{zhai2023sigmoid, tschannen2025siglip}, and PE-Core~\citep{bolya2025perception}, and find that multimodal encoders naturally separate real and AI-generated images in their embedding space. In contrast, this separation is considerably weaker in models trained solely through self-supervision on image datasets (e.g., DINO-series~\citep{caron2021emerging, oquab2023dinov2, simeoni2025dinov3}). Interestingly, beyond the real/fake boundary, embeddings from different generators organize into a small number of higher-level clusters. Generators with similar visual characteristics tend to occupy nearby regions in the embedding space, suggesting that multimodal representations capture shared generative patterns across models. Motivated by this observation, we develop a representation-aware data curation strategy that selects a compact set of representative generators, reducing the training set from 50K to 10K images and from 28 generators to only 8 while preserving broad distributional coverage.


Building on this insight, we propose a lightweight and data-efficient detection framework by attaching a linear classifier to a frozen PE-Core encoder. Furthermore, we construct RealWorldBench, a new test benchmark containing smartphone-captured photos and synthetic images from modern commercial generators.


Using only 10K curated samples—just 1/400 of OpenFake (4M) and 1/30 of AIGIBench (288K)—our PE-Core + Linear classifier achieves state-of-the-art performance on AIGI-Holmes, OpenFake, and RealWorldBench. Moreover, our representation-aware curation strategy consistently outperforms random sampling under the same data budget (96.4\% vs. 94.9\% on RealWorldBench), demonstrating that carefully selected training data can improve robustness to distribution shifts while requiring substantially less data. These results suggest that reliable AI-generated image detection is driven not by larger models or larger datasets, but by robust visual representations and effective data curation.

\vspace{0.1in}
\noindent
We highlight the main contributions of this paper below:
\begin{itemize}[leftmargin=*,itemsep=0mm]
    \item We show that frozen multimodal encoders naturally separate real and AI-generated images while revealing higher-level generator structures.
    \item We propose a representation-aware data curation strategy and a lightweight PE-Core + Linear detector for data-efficient and robust AI-generated image detection.
    \item We introduce RealWorldBench, a test benchmark consisting of authentic photographs and modern commercial synthetic images.
\end{itemize}

\vspace{15pt}


\section{Related Works}
\label{sec:related_works}
\noindent
{\bf AI-generated image detection.}
Early deepfake and synthetic-image detectors primarily relied on CNN-based classifiers trained on specific generative models, such as ProGAN or Stable Diffusion ~\citep{wang2020cnn, rombach2022high}.  
While effective on in-distribution samples, these methods show severe degradation when evaluated on unseen generators or high-quality diffusion models.   Subsequent works focus on artifact-level cues or frequency analysis~\citep{tao2025sagnet, tan2023learning, tan2024frequency, yan2024sanity}, 
yet their reliance on generator-specific traces limits generalization as modern T2I systems become increasingly realistic.  
Recent benchmarks such as AIGIBench~\citep{li2025artificial}, AIGI-Holmes~\citep{zhou2025aigi}, 
and OpenFake~\citep{livernoche2025openfake} introduce large-scale datasets covering a wide range of generators to evaluate cross-generator detection and improve generalization through diverse training data.  
However, these benchmarks rely on extensive data accumulation—requiring large storage, long training time, and continual updates as new T2I models emerge—which makes them difficult to scale in practice.

\vspace{0.05in}
\noindent
{\bf Vision encoders for AI-generated image detection.}
Multimodal vision encoders such as CLIP~\citep{radford2021learning}, SigLIP~\citep{zhai2023sigmoid, tschannen2025siglip}, 
and MLLM-based vision towers (e.g., LLaVA~\citep{liu2023visual}, Qwen-VL~\citep{bai2023qwenvlversatilevisionlanguagemodel, wang2024qwen2}) 
provide strong semantic representations that have recently been explored for synthetic image detection.  
UniFD~\citep{ojha2023towards} and C2C-Clip~\citep{tan2025c2p} demonstrates that linear probes and lightweight adapters over CLIP features improve cross-generator robustness, but the method does not fully exploit 
the inherent generalizability of multimodal embeddings.  
AIGI-Holmes~\citep{zhou2025aigi} takes a different direction by fine-tuning MLLMs via LoRA in 
a two-stage SFT+DPO pipeline to obtain explainable detectors, which requires a large amount of 
carefully annotated preference data.  
Despite these advances, most multimodal approaches still rely on heavy training or explicit 
text–image alignment, rather than leveraging the intrinsic structure already present in 
pretrained vision encoders.
\vspace{-0.02in}
\section{Method}
\begin{figure}
\centering\small
\includegraphics[width=0.98\linewidth]{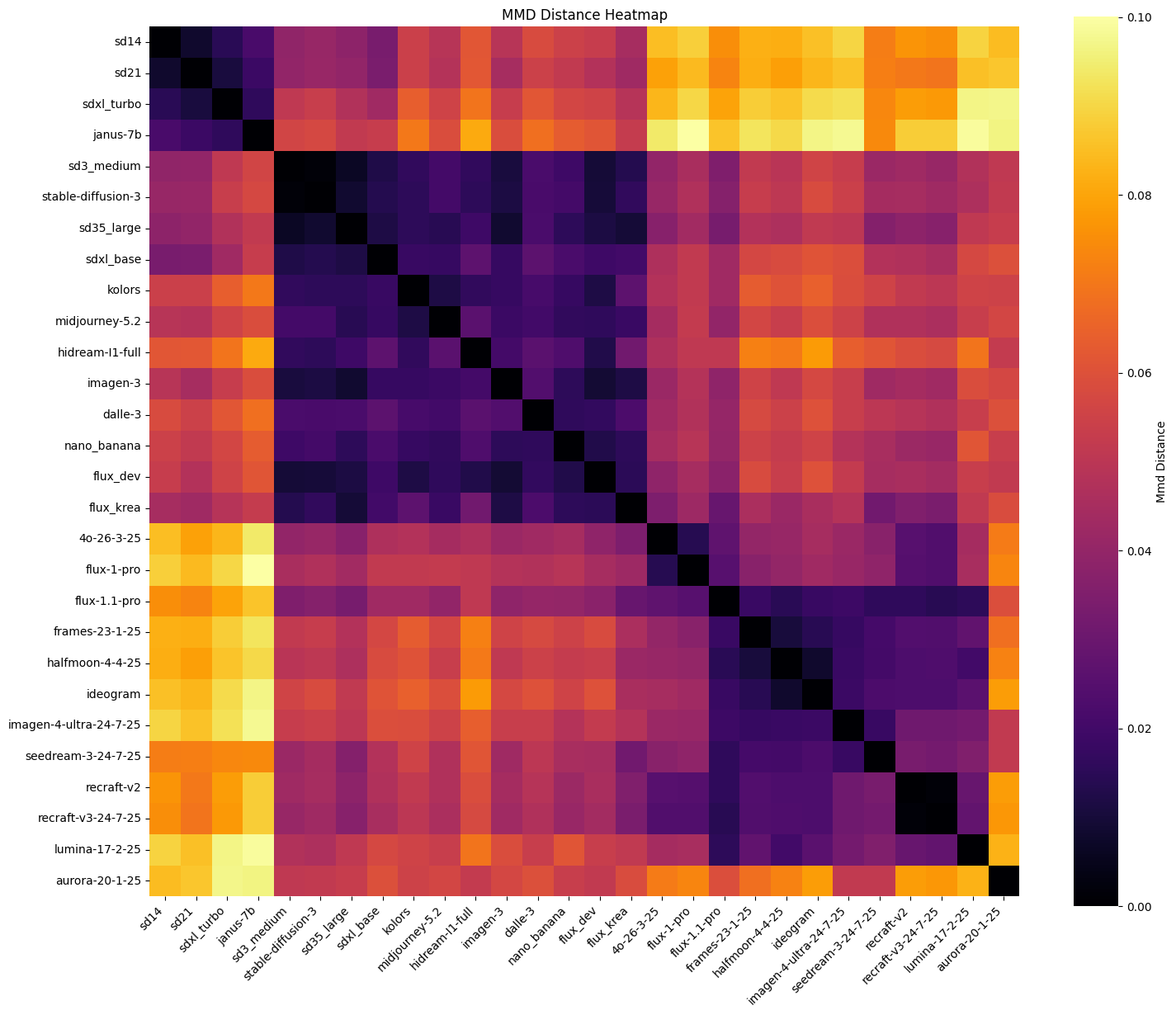}
\vspace{-0.03in}
\caption{
\textbf{Distributional differences across generators in RealWorldBench.}
MMD distances computed in the PE-Core embedding space reveal substantial variation among recent T2I generators, 
highlighting the need for generator-diverse curation and evaluation.
}
\label{fig:realworl-mmd}
\end{figure}

We introduce a simple baseline for detecting AI-generated images using the frozen vision encoder of the multi-modal pre-trained VLMs. Our approach centers on curating a representation-aware dataset and performs linear probing with the embeddings obtained from the frozen encoders on this curated data. Throughout the section, we first elaborate the problem setup (Section~\ref{sec:prob_setup}), and illustrate the motivation (Section~\ref{sec:motivation}) of our approach.
Then, we present an analysis based on the observation using vision encoders to analyze the images generated from various T2I models (Section~\ref{sec:observation}).
Finally, based on this motivation, we propose a novel data curation algorithm (Section~\ref{sec:ds_curation}) and provide details on how to build a simple baseline by just using linear probing on the embeddings (Section~\ref{sec:linear_probe}).

\subsection{Problem Setup}\label{sec:prob_setup}
Our goal is to train a classifier that discriminates the real and fake images. 
Formally, let $h : \mathcal{X} \rightarrow [0,1]$ be a classifier that assigns each image $x \in \mathcal{X}$ to a probability 
indicating that measures how likely it is to be AI-generated
(i.e., it is fully synthesized by a text-to-image model, excluding edited or inpainted content).
Remark that it is a binary classification problem where $h(x)=0$ denotes $x$ is a real image and $h(x)=1$ denotes $x$ is a fake image. 
During inference, we predict the real and fake via thresholding, i.e., 
$\hat{y} =\mathbbm{1}[\,h(x) \ge 0.5\,]$.
To obtain $h$, we consider a hypothesis space $\mathcal{H}$ consisting of linear classifiers, and optimize the classifier via following:
\begin{equation}
    h^{*} = \arg\max_{h \in \mathcal{H}} \; \mathcal{J}(h; \mathcal{D}),
\end{equation}
where $\mathcal{J}$ is a performance measure (e.g., validation accuracy) 
and $\mathcal{D}$ is a curated training dataset.

\subsection{Motivation}\label{sec:motivation}

Recent studies in AI-generated image detection often rely on expanding training datasets by aggregating previous benchmarks or synthesizing large volumes of new images. For example, AIGIBench is based on ProGan and SD1.4; AIGI-Holmes integrates 45,000 images drawn from CNNDetection, GenImage, and DRCT; and OpenFake proposes a massive 4 million-image dataset (approximately 1TB in size). This trend reflects the belief that generalization can be improved simply by accumulating more generators and more fake images. However, this \textit{massive accumulation} approach has several critical drawbacks:
\vspace{5pt}
\begin{enumerate}
    \item \textbf{Unbounded Growth}: Dataset size grows without bound as new T2I generative models continually emerge.
    \item \textbf{Computational Cost}: Training on such large datasets becomes computationally expensive and practically infeasible, since it requires continual updates.
    \item \textbf{Domain Gap}: Existing real-image sources (e.g., ImageNet, LAION, LSUN) are often outdated. They do not reflect modern, high-resolution real images collected from current web platforms or captured by recent smartphone and camera sensors, resulting in a substantial domain gap between training data and real-world deployment scenarios, e.g., current detectors may confuse whether the high-resolution real images is fake or not.
\end{enumerate}
\vspace{5pt}
This indicates that the model architecture is not the core issue; rather, it is the quality and diversity of the training data. This strongly motivates a shift from \textit{more data} to \textit{representation-aware}, \textit{compact}, and \textit{diverse} data curation.

\subsection{Observation Study}\label{sec:observation}

Our approach is grounded in the hypothesis that multi-modal pre-trained vision encoders (e.g., PE-Core, CLIP, and SigLIP) inherently encode both semantic information and subtle \textit{fakeness} cues. To investigate this hypothesis, we compare the embedding spaces of multi-modal encoders and self-supervised encoders (e.g., DINOv2 and DINOv3). Specifically, synthetic images are collected from more than 28 T2I models using the Rapidata Human Preference Dataset~\citep{rapidata_recraft_v3_2025} and other open-source generators, while corresponding real images are retrieved from the same prompt pool using \textit{Pexels}\footnote{\scriptsize\url{https://pexels.com}} and \textit{iStockPhoto}\footnote{\scriptsize\url{https://istockphoto.com/}} to control for semantic content. All analyses are conducted on frozen visual embeddings without using text encoders or captions, isolating generation-induced artifacts from semantic and text-alignment effects. As shown in Figure~\ref{fig:tsne_3}, our analysis revealed two notable properties of multimodal encoders, which are largely absent in self-supervised encoders:
\vspace{5pt}
\begin{itemize}
    \item Natural separation between real and synthetic images in the embedding space.
    \item Higher-level clusters among generative models.
\end{itemize}
\vspace{5pt}




Among all encoders, PE-Core-G14-448 provides the clearest real/fake separation and the most structured generator clusters, and is therefore adopted as the backbone for dataset curation, denoted as \textit{PE-Core}. These observations further motivate our representation-aware curation strategy, which selects representative generators from each cluster to maximize distributional coverage while minimizing redundancy.

\subsection{Dataset Curation}\label{sec:ds_curation}

To construct a compact yet generator-diverse training set, we first estimate the distributional similarity among generators in the embedding space of a frozen vision encoder. Based on these generator-level distances, we group similar generators into hyper-clusters and select representative generators from each cluster. This procedure reduces redundancy among generators while preserving broad coverage over different synthetic image distributions.

\vspace{0.1in}
\noindent
{\bf Step 1: Embedding Extraction.}
Each real or synthetic image is fed into the encoder to obtain an L2-normalized embedding 
$\mathbf{f} \in \mathbb{R}^{D}$.  
These embeddings encode both semantic content and generator-specific \textit{visual fingerprints}, 
making them suitable for distribution-level comparison.

\vspace{0.1in}
\noindent
{\bf Step 2: Generator-Level Distance Estimation via MMD.}
Let $\mathcal{G}=\{g_1,\dots,g_G\}$ be the set of generators, and let
$\mathcal{F}_{g}=\{\mathbf{f}_{g,1},\dots,\mathbf{f}_{g,N_g}\}$ denote the embedding set of generator $g$.
We compute pairwise generator distances using Maximum Mean Discrepancy (MMD):
\begin{align}
    M_{ij}
    &= \mathrm{MMD}(\mathcal{F}{g_i}, \mathcal{F}{g_j}) \nonumber \\
    &= \mathbb{E}\left[k(\mathbf{f},\mathbf{f}')\right]
    + \mathbb{E}\left[k(\mathbf{h},\mathbf{h}')\right]
    - 2\mathbb{E}\left[k(\mathbf{f},\mathbf{h})\right],
\end{align}

where $\mathbf{f}$ and $\mathbf{f}'$ are independently sampled embeddings from $\mathcal{F}{g_i}$,
$\mathbf{h}$ and $\mathbf{h}'$ are independently sampled embeddings from $\mathcal{F}{g_j}$,
and $k(\cdot,\cdot)$ is a Gaussian RBF kernel.
This yields a generator distance matrix $\mathbf{M}\in\mathbb{R}^{G\times G}$.
We then apply hierarchical agglomerative clustering with average linkage on $\mathbf{M}$ to obtain hyper-clusters
$\mathcal{C}={C_1,\dots,C_K}$.

\vspace{0.1in}
\noindent
{\bf Step 3: Representative Generator Selection.}
For each hyper-cluster $C_k$, we select $m$ representative generators.
We compute the centroid embedding of each generator as
\begin{equation}
    \boldsymbol{\mu}_{g} = \frac{1}{|\mathcal{F}_g|} \sum_{\mathbf{f} \in \mathcal{F}_g} \mathbf{f}.
\end{equation}

Representative generators are selected to maximize diversity among generator centroids:
\begin{equation}
\mathcal{R}_k^{*} =
\arg\max_{\mathcal{R} \subseteq C_k, \, |\mathcal{R}|=m}
\min_{g_i, g_j \in \mathcal{R}}
\|\boldsymbol{\mu}_{g_i} - \boldsymbol{\mu}_{g_j}\|_2.
\end{equation}

Since exact optimization is NP-hard, we employ greedy Farthest-Point Sampling (FPS) as an efficient approximation.

\vspace{0.1in}
\noindent
{\bf Step 4: Intra-Generator Sample Selection.}
For each selected generator $g\in\mathcal{R}{k}^{}$, we randomly sample $n_g$ images with distinct prompts:
\begin{equation}
    \mathcal{S}{g}
    =
    \mathrm{RandomSample}(\mathcal{F}{g}, n_g).
\end{equation}
The final curated dataset is
\begin{equation}
\mathcal{D}_{\text{curated}} = \bigcup_{k=1}^{K} \bigcup_{g \in \mathcal{R}_k} \mathcal{S}_g.
\end{equation}

\subsection{Linear Probing with Frozen Encoder}\label{sec:linear_probe}

After constructing the curated dataset $\mathcal{D}_{\text{curated}}$,  
we train a lightweight {linear classifier} on top of frozen PE-Core embeddings.  
Given an image $x$, we first compute its embedding $\mathbf{f} = E(x)$ and then apply a  
single linear layer followed by a sigmoid activation to produce the prediction:
\begin{equation}
    h(x) = \sigma(\mathbf{w}^\top \mathbf{f} + b) \in [0,1],
\end{equation}
where we predict {fake} if $h(x) \ge 0.5$ and {real} otherwise.

Since the encoder remains fixed, training reduces to convex optimization over $(\mathbf{w}, b)$,  
making it extremely efficient while avoiding overfitting to specific generators.  
We optimize the classifier using AdamW with binary cross-entropy loss. No text encoder, prompts, or backbone fine-tuning are required.
This linear probing deliberately emphasizes data quality rather than model complexity, since as we show in our experiments, a linear head trained on our representation-aware curated dataset  
outperforms state-of-the-art deepfake detectors and generalizes robustly across unseen domains.
\vspace{0.1in}

\section{Experiments}
\label{sec:exp}
\begin{figure*}
\centering
\includegraphics[width=0.9\linewidth]{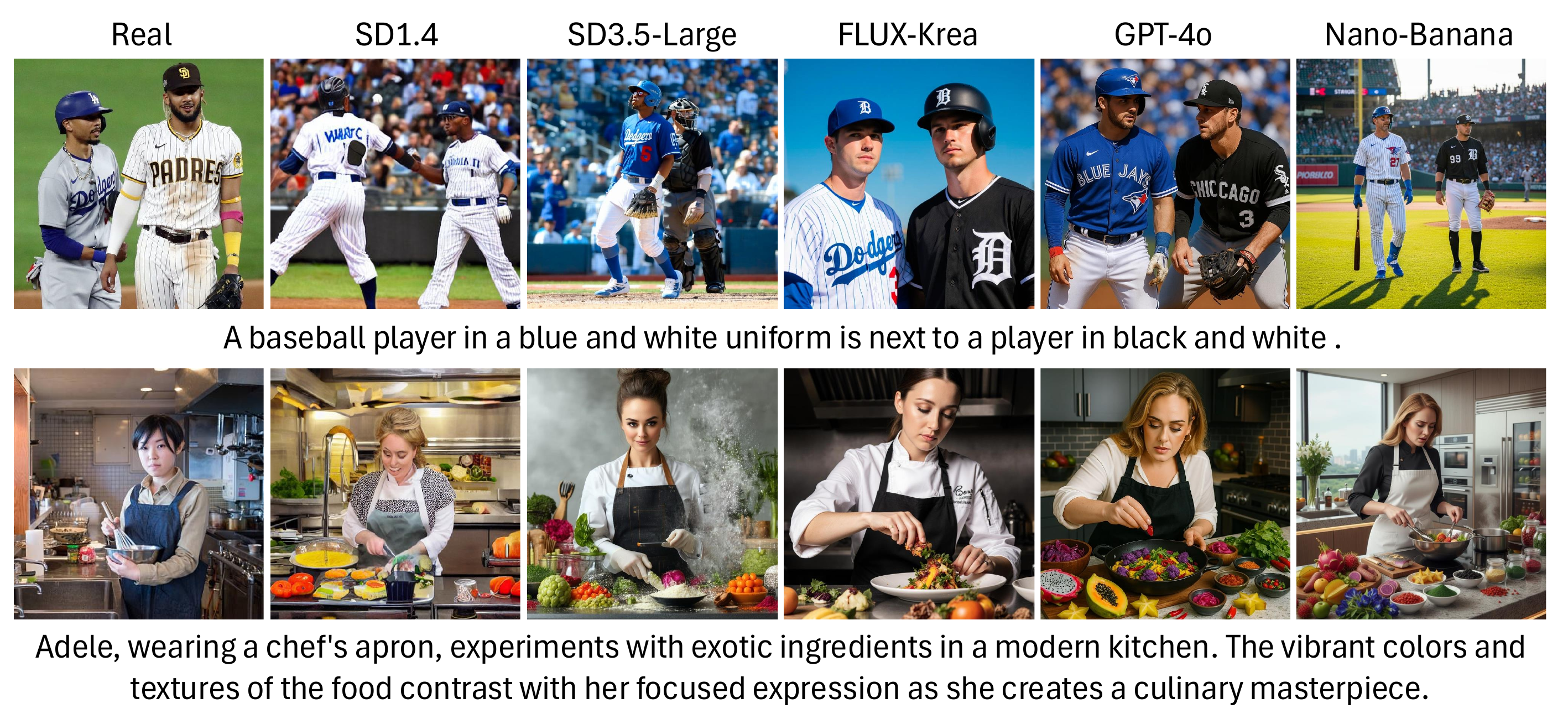}
\vspace{-0.03in}
\caption{\textbf{Examples  from RealWorldBench.} We leverage the T2I human-preference dataset to collect images generated by various commercial models. Additionally, we synthesize more samples using open-source generative models with the same captions as in the preference dataset to preserve the overall image distribution.}
\vspace{-0.03in}
\label{fig:realworldbench_images}
\end{figure*}

\paragraph{Overview.}
Our experiments aims to investigate:
\begin{enumerate}[leftmargin=*]
    \item We evaluate whether a simple PE-Core linear classifier can outperform existing detectors 
    when trained on conventional benchmarks.
    \item We first construct a \textit{universal training set} that spans the full distribution of real images
    (from older low-resolution datasets to modern high-quality smartphone photos) and synthetic images
    (from outdated GANs to recent commercial T2I generative models).
    \item We assess the effectiveness of our \textit{generator-aware dataset curation} when applied to this universal dataset.
    \item We compare the generalization performance of the curated small dataset with the full universal dataset.
    \item We analyze the influence of the vision encoder in two independent stages, i.e., data curation and linear probing, by examining how different encoders affect each stage.
\end{enumerate}
\vspace{5pt}

\noindent
{\bf Common Setup.}
For the benchmarks, we evaluate across four major benchmarks: AIGIBench~\citep{li2025artificial}, AIGI-Homes~\citep{zhou2025aigi}, OpenFake~\citep{livernoche2025openfake}, and our proposed RealWorldBench. For linear probing, we train a one-layer linear classifier on a top of frozen PE-Core embeddings extracted from the curated dataset. Specifically, we use AdamW with learning rate of $1\times10^{-3}$ and batch size of 40.



\subsection{Comparisons with Existing Detection Methods}

\begin{figure}[t!]
\centering\small
\vspace{-0.05in}
\captionof{table}{
\textbf{Baseline detector comparison on AIGIBench.}
Real accuracy (TNR), fake accuracy (TPR), overall accuracy, and average precision (AP) are reported.
All methods are trained on the full AIGIBench training set (288K images), except \textbf{PE-Linear (Curated)}, which is trained on only 10K curated images.
Despite using nearly $29\times$ fewer training samples, PE-Linear (Curated) achieves the best performance across all metrics, indicating that data quality can be more important than dataset scale.
}

\vspace{-0.05in}
\setlength\tabcolsep{3pt}
\begin{tabular}{lcccc}
\toprule
\textbf{Method} & Real Acc. & Fake Acc. & Acc. & A.P. \\
\midrule
CNNDetection \citep{wang2020cnn} & 98.2 & 11.6 & 54.9 & 67.0 \\
Gram-Net \citep{liu2020global} & 90.5 & 26.6 & 58.6 & 62.4 \\
LGrad \citep{tan2023learning} & 85.8 & 39.6 & 62.9 & 66.6 \\
UniFD \citep{ojha2023towards} & 73.3 & 71.5 & 72.5 & 75.6 \\
FreqNet \citep{tan2024frequency}  & 65.9 & 66.4 & 66.2 & 70.1 \\
NPR \citep{tan2024rethinking} & 93.8 & 41.9 & 67.9 & 73.9 \\
Ladeda \citep{cavia2024real} & 91.7 & 54.9 & 73.4 & 79.3 \\
DFFreq \citep{yan2025generalizable} & 91.8 & 58.0 & 75.1 & 82.2 \\
C2P-CLIP* \citep{tan2025c2p} & 93.8 & 49.8 & 71.8 & 82.2 \\
AIDE \citep{yan2024sanity} & 88.1 & 67.0 & 77.6 & 82.7 \\
SAFE \citep{li2024improving} & 89.0 & 66.6 & 78.1 & 83.6 \\
VIB-Net \citep{zhang2025towards}& 60.6 & 78.1 & 69.3 & 70.9 \\
$D^3$ \citep{yang2025d} & 81.0 & 46.4 & 63.7 & 68.9 \\
Effort \citep{yan2024orthogonal} & 96.9 & 57.1 & 77.1 & 87.2 \\
FerretNet \citep{liang2025ferretnet} & 96.6 & 61.8 & 79.4 & 85.8 \\
LOTA \citep{wang2025lota} & 89.3& 65.1& 77.4&83.1\\
\rowcolor{gray!15}
PE-Linear & \underline{98.0} & \underline{78.8} & \underline{88.5} & \underline{95.3} \\
\rowcolor{gray!15}
PE-Linear \textcolor{blue}{(Curated)} & \textbf{98.6} & \textbf{80.1} & \textbf{89.4} & \textbf{95.7} \\
\bottomrule
\end{tabular}

\label{tab:aigibench-baseline}
\end{figure}

\paragraph{Overall Comparison on AIGIBench \citep{li2025artificial}.}
Table~\ref{tab:aigibench-baseline} summarizes the performance of existing deepfake detectors on AIGIBench. All baseline methods are trained under AIGIBench Setting-II using the full training set of 288K images. Despite its simplicity, PE-Linear achieves state-of-the-art performance across all metrics, reaching 88.5\% accuracy and 95.3\% AP. Furthermore, PE-Linear trained on our curated 10K subset improves performance to 89.4\% accuracy and 95.7\% AP while using nearly $29\times$ fewer training samples. This improvement can be explained by a limitation of the original AIGIBench training set: when PE-Core is trained solely on SD1.4 and ProGAN, it achieves near-perfect performance on the in-distribution AIGIBench benchmark. Still, it degrades noticeably on real-world datasets such as CommunityAI and SocialRF (see Appendix~\ref{appendix_c}, Table \ref{tab:aigibench_curation}). To address this gap, our curation strategy supplements the training data with contemporary camera-captured real images and synthetic images from generators not covered by AIGIBench, resulting in broader distributional coverage and improved generalization despite the substantially smaller dataset size. We describe the train set construction and curation procedure in detail in Section~\ref{exp_train_curation}.

\vspace{-10pt}
\paragraph{Overall Comparison on AIGI-Holmes \citep{zhou2025aigi} Benchmark.}

\begin{table*}[ht!]
    \caption{
    \textbf{Cross-generator generalization on the AIGI-Holmes benchmark.}
    Models are trained on 45K images aggregated from existing large-scale GAN and diffusion datasets.
    We evaluate accuracy and AP across 10 different text-to-image generators.
    Our \textbf{PE-Linear} achieves nearly perfect generalization across all generators.
    }
    \label{tab:holmes_baseline}
    \vspace{-4pt}
    \centering
    \renewcommand\arraystretch{1.0}
    \resizebox{0.78\linewidth}{!}{
        \begin{tabular}{lcc|cc|cc|cc|cc|cc}
        \toprule
        \bf Test Dataset $\rightarrow$  & \multicolumn{2}{c}{Janus} & \multicolumn{2}{c}{Janus-Pro-1B} & \multicolumn{2}{c}{Janus-Pro-7B} & \multicolumn{2}{c}{Show-o}& \multicolumn{2}{c}{LlamaGen}& \multicolumn{2}{c}{Infinity}        \\
        \cmidrule{2-13}
        \bf  Detectors $\downarrow$     & Acc.         & A.P.        & Acc.         & A.P.        & Acc.          & A.P.           & Acc.         & A.P.             & Acc.         &A.P.        & Acc.          & A.P.          \\ \toprule
         CNNSpot& 70.0& 86.0& 70.9& 85.8& 85.0& 93.6 & 72.2& 86.0& 61.9&71.4& 86.8& 94.6\\
         AntifakePrompt& 72.2& 87.4& 84.3& 94.0& 84.8& 93.1 & 86.2& 95.5& 96.2&99.4& 83.6& 94.1\\
         UnivFD& 87.6& 97.8& 96.9& 99.5& 96.4& 99.5 & 85.9& 97.4& 93.1&98.6& 79.2& 96.2\\
         NPR& 51.2& 55.9& 69.5& 75.1& 73.9& 77.9 & 93.7& 99.6& 93.5&99.4& 93.8& 99.9\\
         LaRE& 70.8& 99.3& 74.7& 97.5& 95.6& 99.7 & 80.0& 99.0& 91.6&99.6& 77.9& 99.6\\
         RINE& 89.9& 98.3& 98.7& 99.9& 97.2& 99.6 & 98.8& 99.9& 99.1&100.0& 99.2& 99.9\\
         AIDE& 91.2& 99.1& 98.9& 99.9& 97.8& 99.8 & 98.0& 99.8& 99.4&100.0& 98.7& 99.9\\
         AIGI-Holmes& 97.3& 99.9& 99.0& 99.9& 98.0& 99.9 & 99.8& 99.9& 99.9&100.0& 99.9& 100.0\\
        
        \rowcolor{gray!15} PE-Linear &\textbf{ 99.9}& \textbf{100.0}& \textbf{99.9}& \textbf{100.0}& \textbf{99.9}& \textbf{100.0} & \textbf{99.9}& \textbf{100.0}& \textbf{99.9}&\textbf{100.0}& \textbf{100.0}& \textbf{100.0}\\
        \bottomrule
        \end{tabular}}

    \resizebox{0.78\linewidth}{!}{
        \begin{tabular}{
            lcc|cc|cc|cc
             |>{\centering\arraybackslash}p{1.87cm}
            >{\centering\arraybackslash}p{1.87cm}
        }
        \toprule
        \bf Test Dataset $\rightarrow$          &  \multicolumn{2}{c}{VAR}& \multicolumn{2}{c}{PixArt-XL}&\multicolumn{2}{c}{SD3.5-Large} & \multicolumn{2}{c|}{FLUX} & \multicolumn{2}{c}{Mean} \\
        \cmidrule{2-11}
        \bf Detectors $\downarrow$ &  Acc.         & A.P.        & Acc.          & A.P.          &Acc. & A.P. & Acc. & A.P. & Acc. & A.P. \\ \toprule
        CNNSpot &  59.9& 75.0& 78.2& 90.1&63.8 & 81.1 & 79.9 & 92.0 & 72.9 & 85.6 \\
        AntifakePrompt &  90.7& 95.6& 81.7& 92.8&92.8 & 97.8 & 66.1 & 80.8 & 83.9 & 93.1 \\
        UnivFD &  64.3& 85.9& 75.7& 94.4&87.8 & 97.8 & 69.6 & 91.4 & 83.6 & 95.9 \\
        NPR &  85.9& 91.2& 93.4& 99.1&91.6 & 97.7 & 93.6 & 99.5 & 84.0 & 89.5 \\
        LaRE &  98.8& 100.0& 82.2& 99.7&94.1 & 99.5 & 84.3 & 99.0 & 85.0 & 99.3 \\
        RINE &  85.0& 97.9& 98.9& 99.8&97.8 & 99.7 & 97.1 & 99.7 & 96.2 & 99.5 \\
        AIDE &  93.6& 99.3& 98.6& 99.9&99.4 & 100.0 & 94.4 & 99.5 & 97.0 & 99.7 \\
        AIGI-Holmes &  99.6& 100.0& 99.9& 100.0&99.4 & 99.9 & 98.7 & 99.7 & 99.2 & 99.9 \\
        \rowcolor{gray!15} 
        PE-Linear &  \textbf{99.8}& \textbf{100.0}& \textbf{100.0}& \textbf{100.0}&\textbf{99.8} & \textbf{100.0} & \textbf{99.7} & \textbf{99.9} & \textbf{99.9} & \textbf{100.0} \\
        \bottomrule
        \end{tabular}}
    
\end{table*}

Table~\ref{tab:holmes_baseline} presents the cross-generator generalization results on the 
AIGI-Holmes Benchmark. All models are trained on 45K images constructed from existing 
public AI-generated image detection datasets (CNNDetection, GenImage, and DRCT) and evaluated on 
\emph{frontier} text-to-image models.  
Our PE-Linear achieves \textit{the best performance with near-perfect detection}, 
reaching 99.9\% accuracy and 100.0\% AP across all test generators.
The AIGI-Holmes training set combines ProGAN and LSUN reals from CNNDetection, 
ImageNet reals and synthetic images generated by ADM, BigGAN, GLIDE, Midjourney, SD1.4/1.5, VQDM, and Wukong from GenImage, and a wide collection of diffusion-based manipulations from DRCT. This diverse mixture of GAN and diffusion families makes AIGI-Holmes an ideal starting point for analyzing generator-level structure and for applying our dataset curation.

To further assess robustness, we evaluate the model trained on the AIGI-Holmes dataset on the
AIGIBench test split. The detector retains near-perfect performance across almost all synthetic domains but shows clear weaknesses on \textit{modern high-quality real images}, especially those from CommunityAI and SocialRF.  
We additionally evaluate on the Chameleon dataset \citep{yan2024sanity}, which contains user-created AI-generated images from active online art communities.  
As shown in Table~\ref{tab:holmes-failure}, the model continues to detect synthetic images with high confidence but suffers substantial degradation in real-image accuracy, highlighting a central limitation of training solely on the AIGI-Holmes dataset.

\begin{figure}[t!]
\centering
\vspace{0.05in}
\centering\small
\captionof{table}{
\textbf{Failure cases of PE-Linear trained on AIGI-Holmes.}  
While the model achieves near-perfect fake detection across all scenarios, its performance 
significantly degrades on modern high-quality real images from SocialRF, CommunityAI, and Chameleon, revealing a key limitation of training solely on AIGI-Holmes.
}
\vspace{-0.05in}
\resizebox{\linewidth}{!}{
\begin{tabular}{lcccc}
\toprule
Real-World Scenario & {Real Acc.} & {Fake Acc.} & {Acc.} & {A.P.} \\
\midrule
SocialRF      & 41.8 & 99.5 & 70.7 & 95.3 \\
CommunityAI   & 66.0 & 99.7 & 82.8 & 99.5 \\
Chameleon     & 66.9 & 99.7 & 83.3 & 98.8 \\
\bottomrule
\end{tabular}
\label{tab:holmes-failure}
}
\vspace{-0.1in}
\end{figure}

\paragraph{Performance on OpenFake \citep{livernoche2025openfake} Benchmark.}

\begin{table}[ht!]
    \caption{
    \textbf{OpenFake generalization under data subsampling.}
    We train PE-Linear using subsets of size 5K, 10K, 30K, and 100K from the OpenFake training set.
    The full SwinV2 model trained on all 4M images is shown for reference.
    Our {30K curated subset} surpasses the 4M-image SwinV2 model,
    and even 10K or 5K curated datasets achieve competitive performance.
    Generators highlighted in \textcolor{blue}{blue} are out-of-distribution (OOD) for all detectors.
    }
    \label{tab:exp3_openfakebench}
    \vspace{-3pt}
    \centering
    \renewcommand\arraystretch{1.0}
    \resizebox{0.93\linewidth}{!}{
        \begin{tabular}{lcccc|c}
        \toprule
        Model&    \multicolumn{4}{c}{PE-Linear}&SwinV2\\
 \cmidrule{1-6}
 Num data&   5K&10K&30K&  100K&4M\\ \toprule
 SD 1.5 &  100.0&96.9&100.0&  100.0&100.0\\
 SD 2.1 &  100.0&100.0&100.0&  100.0&100.0\\
 SD XL  &  100.0&100.0&100.0&  100.0&100.0\\
 SD 3.5 &  99.6&100.0&100.0&  99.4&100.0\\
 \cmidrule{1-6}
 Flux 1.0 Dev         &  96.7&99.4&98.8&  99.4&100.0\\
        Flux-1.1-Pro         &  96.5&91.4&97.9&  96.4&100.0\\
 Flux-1.0-Schnell     &  99.1&98.4& 100.0& 100.0& 99.9\\
 \cmidrule{1-6}
 Midjourney 6         &  100.0&100.0& 100.0& 100.0& 100.0\\
 Midjourney 7         &  99.4&100.0& 100.0& 99.4& 99.4\\
 DALL\textperiodcentered E~3 &  100.0&100.0& 100.0& 99.4& 99.5\\
 GPT Image 1          &  99.1&98.7& 98.7& 98.7& 99.8\\
 Ideogram 3.0         &  98.0&98.3& 97.7& 98.3& 100.0\\
 Imagen 3.0           &  98.9&99.3& 98.6& 100.0& 99.9\\
 Imagen 4.0           &  99.2&99.4& 99.4& 100.0& 99.6\\
 Grok 2               &  99.8&98.8& 100.0& 100.0& 100.0\\
 HiDream-I1 Full      &  96.4&97.5& 99.4& 99.4& 100.0\\
 Chroma               &  98.6&98.8& 100.0& 99.4& 99.2\\
 \cmidrule{1-6}
 \textcolor{blue}{\textit{Ideogram 2.0}} &  100&100.0& 100.0& 97.9& 99.3\\
 \textcolor{blue}{\textit{Lumina}}       &  100&100.0& 100.0& 100.0& 100.0\\
 \textcolor{blue}{\textit{Frames}}       &  100&100.0& 100.0& 100.0& 96.8\\
 \textcolor{blue}{\textit{Halfmoon}}     &  99.0&100.0& 100.0& 100.0& 99.5\\
 \textcolor{blue}{\textit{Recraft v2}}   &  100&100.0& 100.0& 100.0& 97.2\\
 \textcolor{blue}{\textit{Recraft v3}}   &  98.7&99.3& 97.2& 91.0& 70.1\\
 \cmidrule{1-6}
 \rowcolor{gray!15} Real  (TNR)&  99.2&99.4&99.5&  \textbf{99.8}&99.5\\
 \cmidrule{1-6}
 \rowcolor{gray!15}Average TPR           &  99.0&98.3& \textbf{99.2}& 99.1& 98.8\\
 \rowcolor{gray!15}{\footnotesize Overall} F1            &  99.0&98.8& 99.3& \textbf{99.5}& 99.2\\
 \rowcolor{gray!15}\small{Overall} ROC AUC&  99.4&99.9& \textbf{100.0}& \textbf{100.0}& \textbf{100.0}\\       
 \rowcolor{gray!15}\small{Overall} PR AUC        & 99.4&99.9&\textbf{100.0}&  \textbf{100.0}&\textbf{100.0}\\
        \bottomrule
    \end{tabular}
}
\end{table}

Table~\ref{tab:exp3_openfakebench} presents the generalization results on the OpenFake benchmark under various training sizes. Although the full OpenFake dataset contains over 4M
images, PE-Linear requires only a tiny fraction of this data. For example, even with 5K samples, the model already performs competitively—representing an 
800× reduction relative to the full dataset.  
With 10K samples, performance further improves.  
Most notably, training with just 30K samples—a 133× reduction—actually surpasses the SwinV2 detector trained on all 4M images.  
These results demonstrate that massive data accumulation is unnecessary for detecting AI-generated images: with our curation strategy, less than 1\% of the original dataset is sufficient to capture the generative diversity of OpenFake and achieve state-of-the-art generalization across both in-distribution and OOD generators.




\subsection{Universal Training Set Construction}\label{exp_dset_construction_train}

To build a training set that accurately reflects modern real--image distribution while maintaining
broad coverage of synthetic image sources, we construct a 50K universal training set
through the following process.

\paragraph{Incorporating Modern High-Resolution Real Images.}
Our analysis of existing benchmarks shows that detectors trained on AIGI-Holmes correctly recognize almost all synthetic images but struggle on modern high-resolution real photographs. This limitation stems from the fact that the AIGI-Holmes dataset contains only LSUN- and ImageNet-style real images, which do not reflect the characteristics of contemporary smartphone or web-crawled photos.
To address this distributional gap, we augment the real images from the subset of AIGI-Holmes with an additional 5K high-quality images collected from Open Images V7 \footnote{\scriptsize\url{storage.googleapis.com/openimages/web/index.html}}, Pexels, and Pixabay\footnote{\scriptsize\url{https://pixabay.com}}, resulting in a more diverse and modern real-image distribution.
For synthetic images, we retain the full set of Holmes fake training data, which already covers a wide spectrum of GAN, diffusion, and transformer-based T2I models. Since prior experiments indicate that AIGI-Holmes provides sufficient synthetic coverage,  we do not expand fake images at this stage.

\paragraph{Universal Training Set.}
Combining the augmented real images (5K) with the original AIGI-Holmes training set (45K) yields a unified 50K universal training set.  
This dataset serves as the initial candidate pool from which we later derive a compact subset using our representation-driven data curation method. In the following section, we apply our curation algorithm to obtain a significantly smaller yet diversity-preserving subset for training an robust AI-generated detector.

\paragraph{Domain Analysis of AIGI-Holmes.}
A closer examination of the AIGI-Holmes training set reveals that its real and synthetic images
originate from heterogeneous semantic sources.  
For instance, real images include COCO photographs, ImageNet classes, and LSUN scenes, while the
corresponding synthetic images are generated using text prompts derived from COCO captions or from
ImageNet class names (e.g., ADM, BigGAN, GLIDE, SD-v1.4/1.5, VQDM, Wukong).  
ProGAN images are also paired with LSUN real images.  
As a result, Holmes contains multiple domain-specific real/fake pairs, each with different
semantic distributions.

\subsection{Train Set Curation}\label{exp_train_curation}
We apply our representation-based curation method (Section~\ref{sec:ds_curation}) to the 50K universal training pool. 
Since AIGI-Holmes spans multiple semantic domains (COCO, ImageNet, LSUN), we curate each domain independently to preserve semantic consistency.

Using PE-Core embeddings, we cluster generators by distribution similarity and retain only representative ones. 
For ImageNet, the original eight generators (e.g., Midjourney, SD1.4/1.5, Wukong, GLIDE, ADM, VQDM, BigGAN) are reduced to four: {SD1.4, BigGAN, GLIDE, Midjourney}. 
For COCO, we retain three key generators: {SDXL, SD1.5, SD2.1}, and additionally include {SD3.5} to capture recent diffusion trends. 
Real images are drawn from a balanced mixture of LSUN, COCO, ImageNet, OpenImages, and high-resolution web photos.

Overall, our procedure compresses the 50K pool into a compact {10K curated set} and reduces the original 28 domain–generator combinations to only {8 representative combinations}, while maintaining semantic alignment and generator-level diversity.

\subsection{Test Set Construction}\label{exp_test_construction}

To address the limitations of existing benchmarks—outdated real images and limited coverage of recent generators—we construct {RealWorldBench}, enabling us to evaluate detector robustness on {modern} real photographs and diverse commercial T2I models.

RealWorldBench includes high-resolution real images from recent smartphones and curated web platforms (e.g., {Pexels}, {Pixabay}), manually filtered to remove AI-generated content.  
It also contains synthetic outputs from 28 recent diffusion, transformer, and commercial generators, such as FLUX-1.1-Pro, Flux-Krea, Imagen-3/4/4-Ultra, DALL\textperiodcentered E~3, Nano-Banana, Seedream~3, and Recraft-v2/v3.  
This combination of modern real photos and OOD synthetic images provides a realistic and challenging testbed for evaluating robustness gains from our curated training set.

\subsection{RealWorldBench Evaluation}
Table~\ref{tab:exp4_realworldbench} shows that detectors trained on AIGIBench and OpenFake generalize poorly to modern real images (65.6\% and 29.7\% TNR) and many OOD generators.  
In contrast, our {Curated 10K} model achieves the best real-image robustness ({98.3\%} TNR) and strong cross-generator performance, closely matching the Universal 50K model despite using 5× fewer samples.  
The results show that representation-guided curation is substantially more effective than scaling dataset size for real-world deepfake detection.

\begin{table}[t!]

    \caption{
    \textbf{Performance on RealWorldBench.}
    Our \textbf{Curated 10K} set achieves the best real-image performance (98.3\% TNR) and competitive OOD generalization with far fewer training samples.
    }
    \vspace{-3pt}
    \centering
    \renewcommand\arraystretch{1.0}
    \resizebox{1.0\linewidth}{!}{
        \begin{tabular}{l cccc}
        \toprule
Train dataset&   AIGIBench&OpenFake&Universal&Curated \\
\midrule
Num Train data&  288K&50K& 50K& 10K\\
Generation Methods&  2&35& 28&8\\
 \toprule
 SD 1.4&  90.2&70.4&98.4&98.3\\
 SD 2.1 &  90.9&74.1&100.0&99.9\\
 SDXL-Turbo&   75.0&70.4&100.0&100.0\\
 SDXL-Base&  89.1&100.0&100.0&100.0\\
 SD3 Medium&  83.8&100.0&100.0&99.9\\
  SD3.5 Large&  68.9&98.7&97.2&99.0\\
        Kolors&  76.3&100.0&100.0&100.0\\
 Flux 1.0 Dev&  56.4&98.8&98.6&99.0\\
 Flux Krea&   46.3&95.8&91.7&91.6\\
 \midrule
 Flux 1.0 Pro&   84.1&99.8&91.7&89.0\\
 Flux 1.1 Pro&  95.2&99.3&92.9&90.8\\
 DALL\textperiodcentered E~3 &  74.0&100.0&100.0&99.9\\
 Imagen 3&  51.4&98.4&96.1&98.6\\
 Imagen 4&  81.1&100.0&95.3&97.1\\
 Imagen 4 Ultra&  99.2&97.1&95.4&94.2\\
 Nano Banana&  41.7&99.0&98.4&98.7\\
 Midjouney 5.2&  62.1&99.2&99.6&99.6\\
 GPT-Image-1&   72.7&100.0&92.7&94.2\\
 Janus 7B&  75.1&64.3&100.0&99.9\\
 Ideogram&  97.3&99.9&81.0&72.4\\
 HiDream-I1 Full      &  47.1&100.0&99.9&99.8\\
 Lumina&  100.0&100.0&100.0&99.3\\
 Frames&  95.1&98.9&81.6&80.8\\
 Halfmoon&  93.3&100.0&79.5&74.8\\
 Aurora&  70.8&100.0&99.8&98.9\\
 Seedream 3&   80.6&98.9&84.6&77.9\\
 Recraft v2&  99.1&98.1&96.7&93.6\\
 Recraft v3&  99.0&97.0&97.0&95.2\\
 \midrule
 \rowcolor{lightgray!20} Real (TNR)&  65.6&29.7&96.8&\textbf{98.3}\\
 \midrule
 \rowcolor{lightgray!20}Average TPR           &  78.4&94.9&\textbf{95.3}&94.4\\
 \rowcolor{lightgray!20}\small{Overall} ROC AUC&  81.5&87.5&\textbf{99.5}&99.0\\       
 \rowcolor{lightgray!20}\small{Overall} PR AUC &  79.3&87.5&\textbf{99.3}&99.0\\
        \bottomrule
    \end{tabular}
}
\label{tab:exp4_realworldbench}
\end{table}

\begin{table}[t]
\centering
\small
\caption{\textbf{Effect of representation-aware data curation.}
Universal uses all 28 available generators. Curated selects 8 representative generators using the proposed embedding-space clustering strategy, whereas Random uses the same generator budget but selects generators uniformly at random. Random results are averaged over ten independent runs to reduce sampling variance.}

\label{tab:curation_ablation}
\resizebox{1.0\linewidth}{!}{

\begin{tabular}{lcccc}
\toprule
\textbf{Train Set} &
\textbf{\#Imgs} &
\textbf{\#Gens} &
\textbf{AIGIBench} &
\textbf{Realworld} \\
\midrule
Universal & 50K & 28 & \textbf{90.3} & 96.0 \\
Random & 10K & 8 & 86.3 & 94.9 \\
\rowcolor{gray!15}
Curated (Ours) & 10K & 8 & \underline{89.4} & \textbf{96.4} \\
\bottomrule
\end{tabular}
}

\end{table}
\paragraph{Effect of Representation-Aware Data Curation.}
Table~\ref{tab:curation_ablation} compares different train set construction strategies. Using the same 10K data budget, Curated consistently outperforms Random on both AIGIBench (89.4\% vs. 86.3\%) and RealworldBench (96.4\% vs. 94.9\%), demonstrating that selecting representative generators is substantially more effective than random sampling. Furthermore, Curated achieves performance comparable to Universal while using only one-fifth of the training samples. The gain is particularly pronounced on RealworldBench, whose test distribution differs substantially from the training data in both image content and generator composition, suggesting that representation-aware curation improves robustness to distribution shifts and strengthens out-of-distribution generalization.

\vspace{0.05in}
\noindent
{\bf Implementation Details.}
All experiments are implemented in PyTorch~2.5 with CUDA~12.4.  
We evaluate on AIGIBench, AIGI-Holmes, OpenFake, and our proposed RealWorldBench.  
All random seeds and dataset splits are fixed for reproducibility.  
Code, curated generator lists, and evaluation scripts will be released upon publication.
\section{Conclusion}


We presented a simple and effective baseline for detecting fully AI-generated images using a frozen multimodal vision encoder and a linear classifier. Our analysis shows that multimodal vision encoders inherently capture both semantic structure and generator-specific cues, enabling strong real--fake separability without task-specific fine-tuning. Building on this observation, we introduced a representation-driven data curation strategy that selects a compact set of distributionally representative generators. This approach reduces the universal training pool from 50K to 10K images while compressing 28 generators into only 8 representative ones. Despite this substantial reduction, the curated training set maintains broad coverage of the synthetic image distribution and improves data efficiency. Across AIGIBench, AIGI-Holmes, OpenFake, and RealWorldBench, our frozen-encoder baseline achieves state-of-the-art performance and strong out-of-distribution generalization while using significantly less training data.

\newpage
{
    \small
    \bibliographystyle{ieeenat_fullname}
    \bibliography{main}
}

\clearpage
\onecolumn
\appendix

\setcounter{page}{1}

\begin{center}
    {\bf {\Large SSAFE: Simple and Strong AI-Generated Image Detection via Frozen Vision \\ \vspace{2mm} Encoders}} \\ 
    \vspace{4mm}
    { \Large Supplementary Materials}
    
\end{center}

\vspace{3mm}

\section{Data Curation Pipeline}
\label{sec:appendix_curation}

This section provides detailed descriptions of how our data curation pipeline is applied to the AIGI-Holmes \citep{zhou2025aigi} dataset in practice. 
While the main paper discusses the overall methodology, here we focus exclusively on the concrete procedure used to derive our curated training set from the original AIGI-Holmes collection and auxiliary real-image sources.

\subsection{Overview of AIGI-Holmes Train Set}  
Our training set comprises roughly 45K images curated from widely used public AI-generated image detection datasets, including CNNDetection~\citep{wang2020cnn}, GenImage~\citep{zhu2023genimage}, and DRCT~\citep{chen2024drct}.

\vspace{0.05in}
\noindent
{\bf Real Images.}
The real-image portion of AIGI-Holmes is sourced primarily from three widely used vision datasets:
\begin{itemize}[leftmargin=10pt]
    \item {LSUN} \citep{yu2015lsun} (car, cat, chair, horse, and indoor scenes)
    \item {ImageNet} \citep{5206848} (object-centric natural photographs)
    \item {COCO 2017} \citep{lin2014microsoft} (complex multi-object scenes)
\end{itemize}

\vspace{0.05in}
\noindent
{\bf AI-Generated Images.}
To ensure consistency and interpretability, we reorganize all synthetic samples into
\textit{generator-centric groups} and when relevant, annotate the approximate \textit{image domains}
(e.g., ImageNet-like or COCO-like distributions) associated with each model family.
A comprehensive summary of all generators included in the AIGI-Holmes training split is provided in Table~\ref{tab:aigi-holmes-statistics}. This domain-based organization provides a clearer understanding of how AIGI-Holmes aggregates heterogeneous generative sources across architecture families and training datasets.

\subsection{Representation Analysis and Hypercluster-Based Selective Sampling}

To characterize the AIGI-Holmes dataset and identify representative samples, we embed every image using a frozen PE-Core-G14-448 encoder. Using the same encoder for both feature extraction and downstream training ensures that the curated dataset aligns with the classifier’s representation space.

\begin{figure*}
\centering
\vspace{-0.3in}

\begin{subfigure}{0.7\linewidth}
    \centering
    \includegraphics[width=\linewidth]{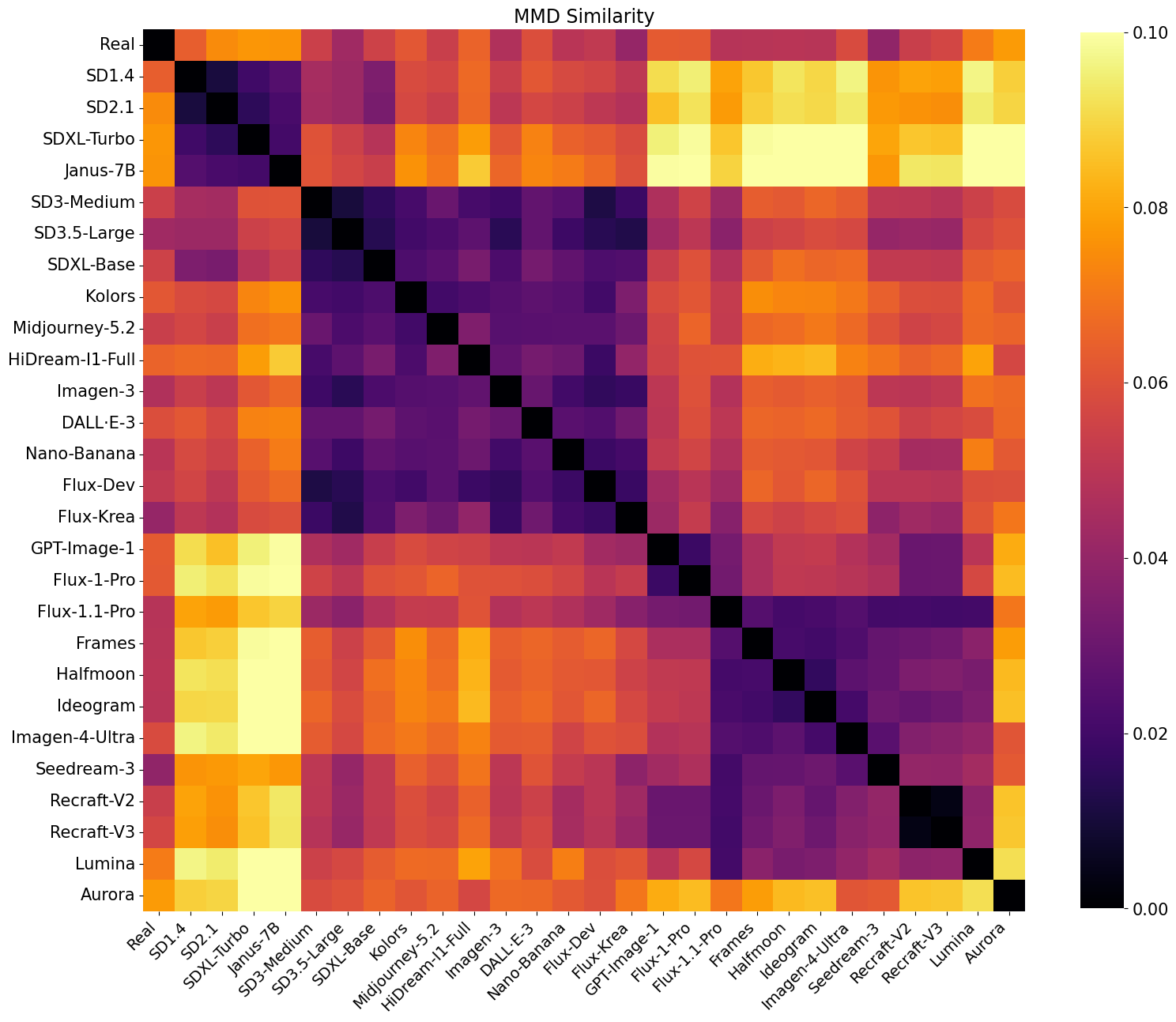}
    \caption{PE-Core-G14-448}
    \label{fig:mmd-pe}
\end{subfigure}

\hfill

\begin{subfigure}{0.7\linewidth}
    \centering
    \includegraphics[width=\linewidth]{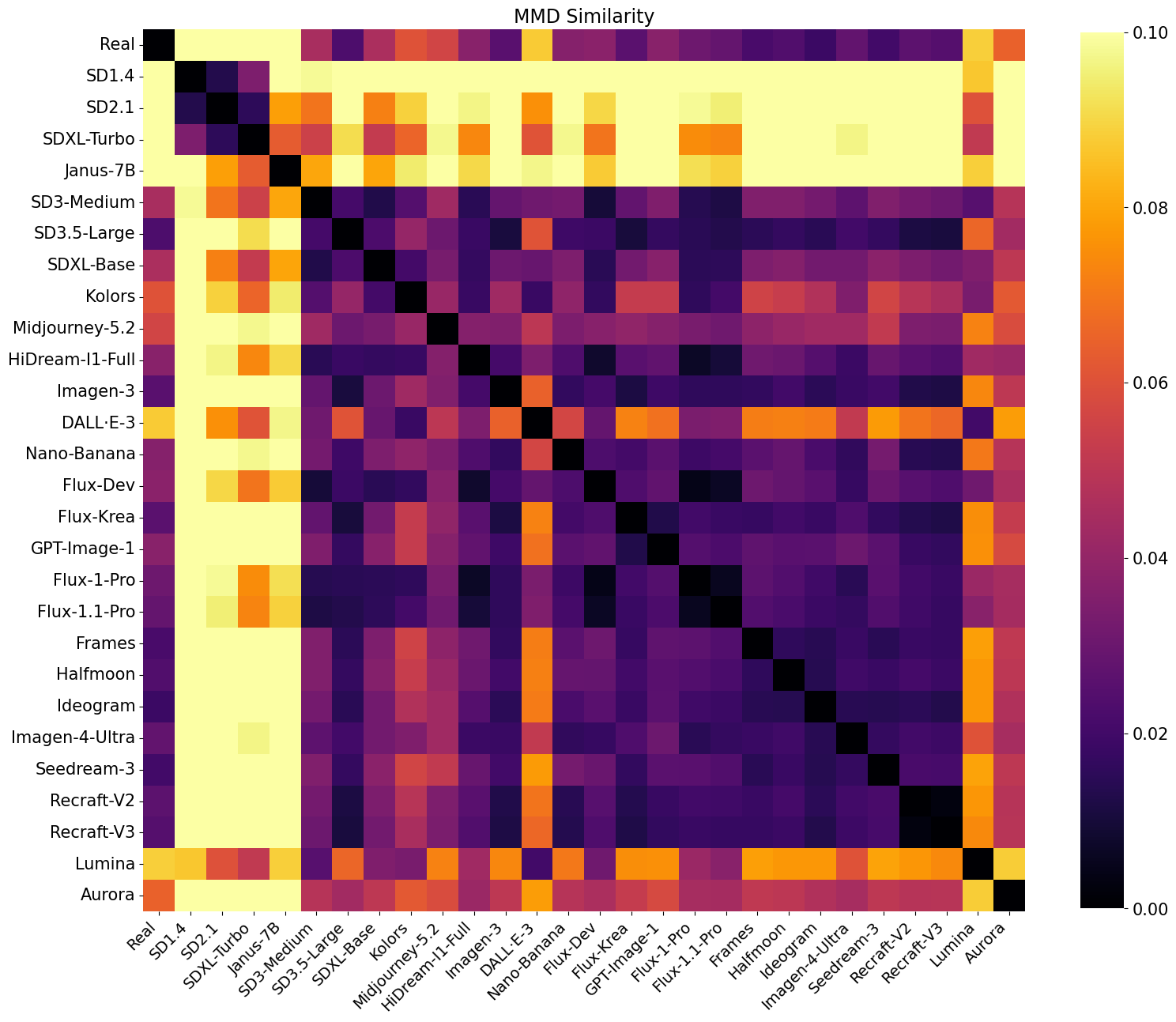}
    \caption{SigLIP2-G16-384}
    \label{fig:mmd-siglip}
\end{subfigure}

\caption{\textbf{MMD structures of multimodally pretrained encoders.}
PE-Core-G14-448 and SigLIP2-G16-384 exhibit clear separation between real and fake images, 
with fake samples forming distinct generator-specific clusters. 
The effect is most pronounced in PE-Core.}
\label{fig:mmd-big-pe-siglip}

\end{figure*}

\begin{figure*}
\centering
\vspace{-0.3in}

\begin{subfigure}{0.7\linewidth}
    \centering
    \includegraphics[width=\linewidth]{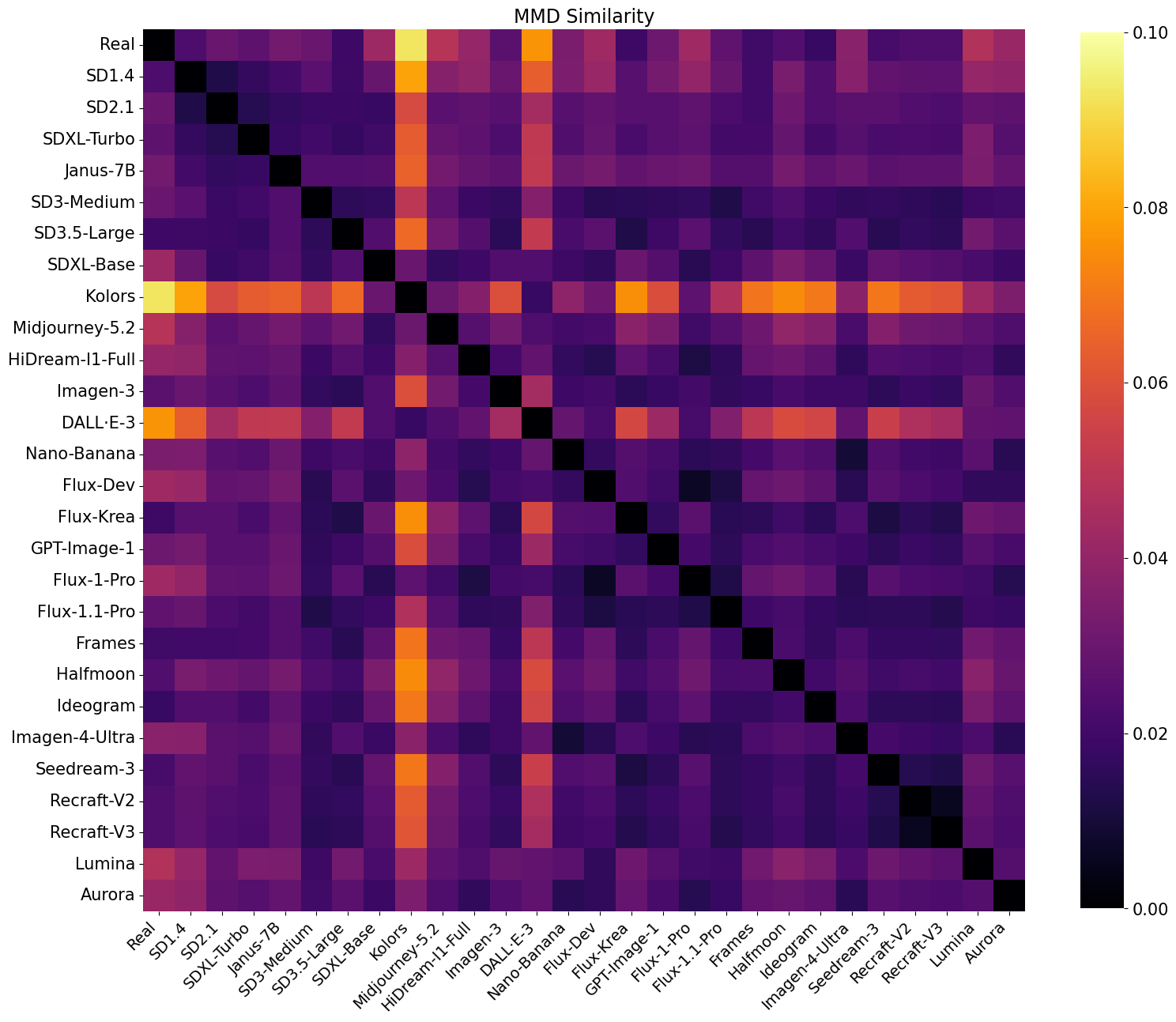}
    \caption{DINOv3-L/16}
    \label{fig:mmd-dinov3}
\end{subfigure}

\hfill

\begin{subfigure}{0.7\linewidth}
    \centering
    \includegraphics[width=\linewidth]{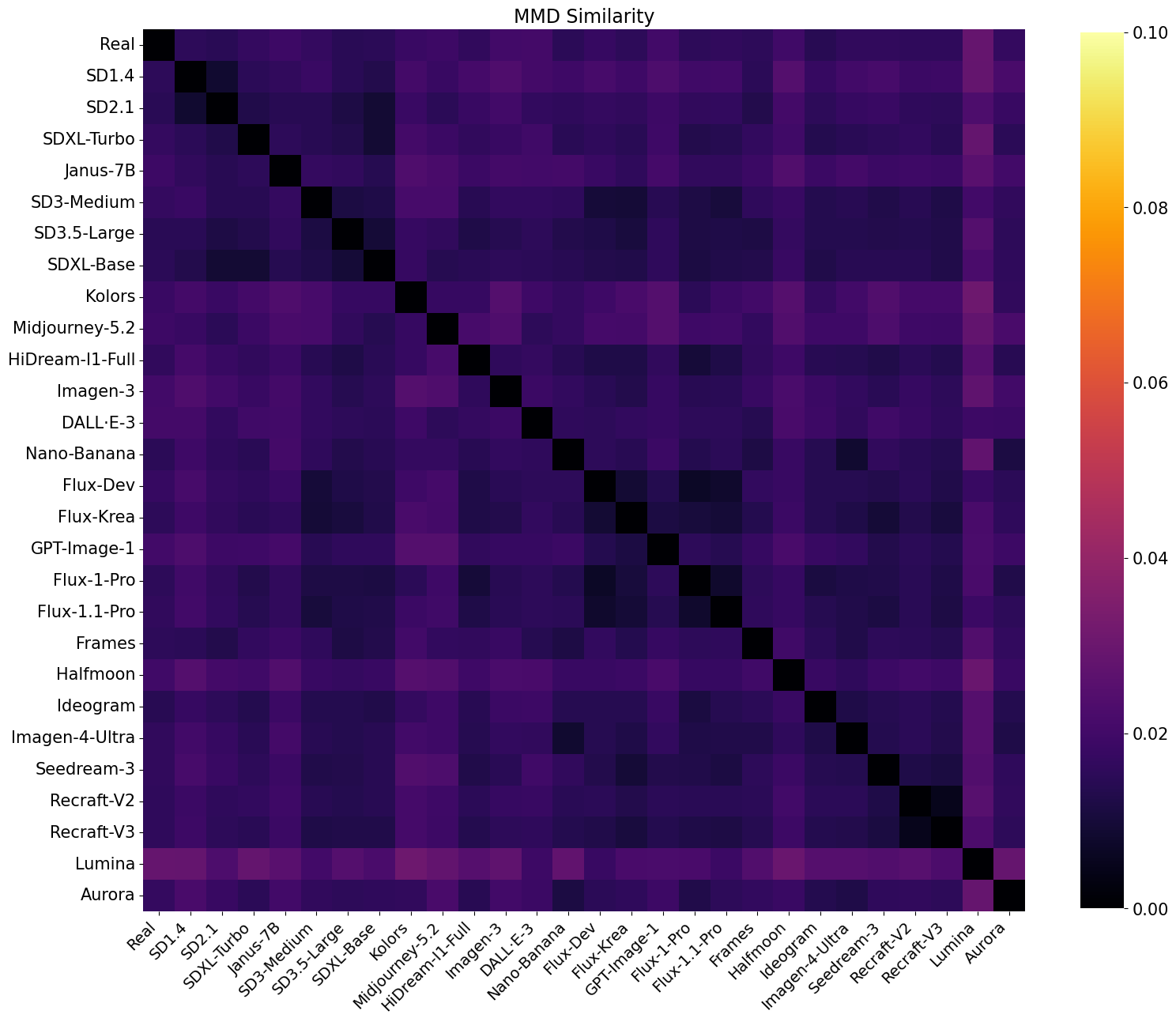}
    \caption{DINOv2-Giant}
    \label{fig:mmd-dinov2}
\end{subfigure}

\caption{\textbf{MMD structures of self-supervised encoders.}
Unlike multimodally pretrained models, DINOv3-L/16 and DINOv2-Giant do not show meaningful separation between real and fake images, indicating that self-supervised representations do not inherently encode generative-model artifacts.}
\label{fig:mmd-big-dino}

\end{figure*}

\vspace{0.05in}
\noindent
{\bf Representation structure across encoders.}
While the generator-level MMD structure highlights how AIGI-Holmes is organized in the PE-Core embedding space, we further examine whether this behavior generalizes across different encoder families. For a fair comparison, we subsample image pairs from RealWorldBench such that each encoder is evaluated on the same set of semantically aligned real–synthetic image groups. Figure~\ref{fig:mmd-big-pe-siglip} and Figure~\ref{fig:mmd-big-dino} present the resulting MMD matrices for multimodally pretrained encoders (PE-Core \citep{bolya2025perception}, SigLIP2 \citep {tschannen2025siglip}) and self-supervised encoders (DINOv3 \citep{simeoni2025dinov3}, DINOv2 \citep{simeoni2025dinov3}). The PE-Core matrix shown in Figure~\ref{fig:mmd-big-pe-siglip} corresponds to a large-scale version of the visualization in Figure~\ref{fig:realworl-mmd}. Multimodal encoders exhibit clear real–fake separation and well-defined generator clusters, providing reliable signals for identifying redundancy. In contrast, self-supervised encoders produce entangled structures where real and synthetic images overlap, indicating that they do not naturally encode generative artifacts. This validates our use of PE-Core for representation-aware curation.

\vspace{0.05in}
\noindent
{\bf Hypercluster discovery and Representative generator selection.}
Using PE-Core embeddings, we compute generator-level MMD distances and apply hierarchical clustering to identify coherent hyperclusters in feature space. These clusters summarize distributional similarities and provide a principled basis for generator selection. For each hypercluster, we retain only a small number of representative generators (typically two) and draw samples uniformly at random from them. The curated dataset does not rely on sample-level filtering; its diversity arises from generator-level selection.

\begin{figure*}
\centering

\begin{subfigure}{0.31\linewidth}
    \centering
    \includegraphics[width=\linewidth]{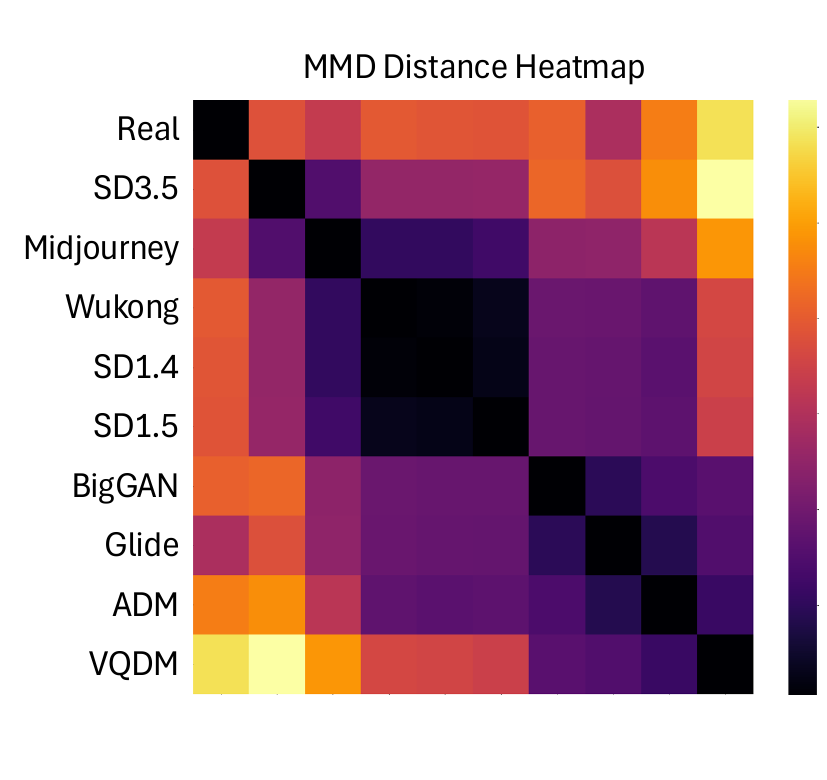}
    \caption{MMD Distance Heatmap}
    \label{fig:mmd-heatmap}
\end{subfigure}
\hfill
\begin{subfigure}{0.35\linewidth}
    \centering
    \includegraphics[width=\linewidth]{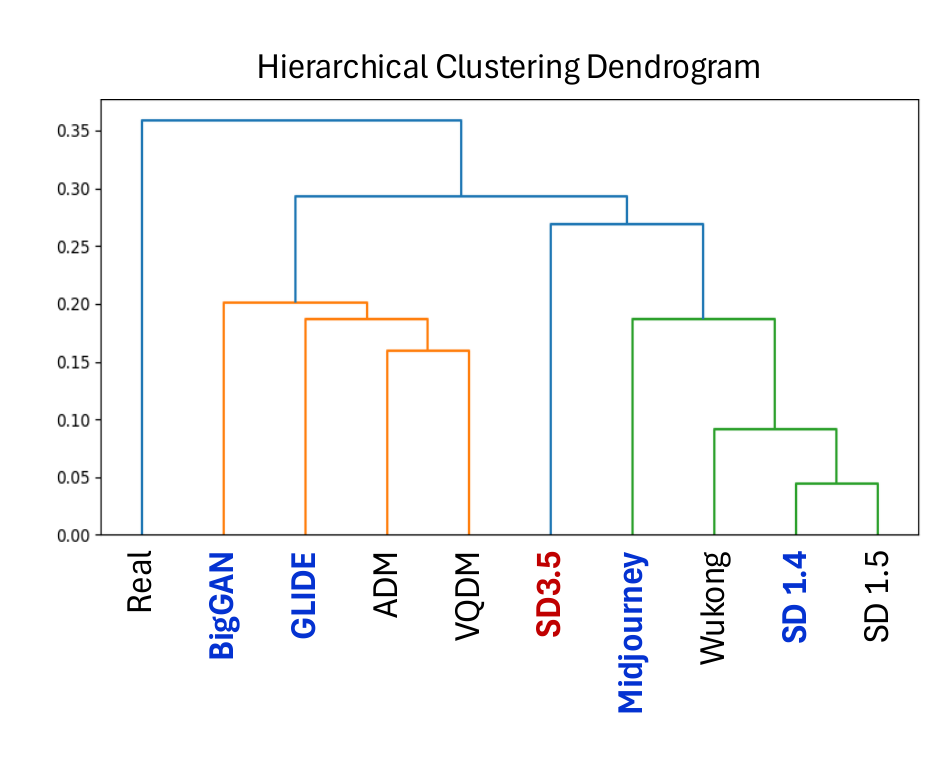}
    \caption{Hierarchical Clustering}
    \label{fig:mmd-dendrogram}
\end{subfigure}
\hfill
\begin{subfigure}{0.3\linewidth}
    \centering
    \includegraphics[width=\linewidth]{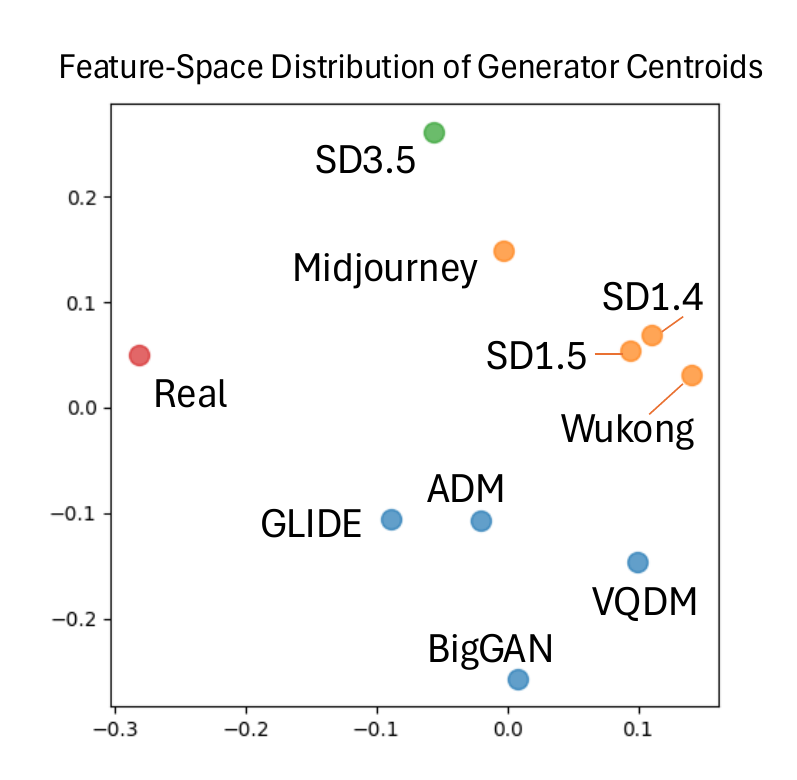}
    \caption{2D Centroid Projection}
    \label{fig:mmd-centroids}
\end{subfigure}

\vspace{-0.05in}
\caption{\textbf{Generator-level similarity analysis using PE-Core embeddings.}
(a) MMD distance matrix between real and synthetic generators,
(b) hierarchical clustering revealing hypercluster structure, and
(c) 2D projection of generator centroids.
The analysis shows that synthetic generators form three major clusters, from which we select four representative domains (\textcolor{blue}{highlighted in blue}).
For \textcolor{red}{SD3.5}, we additionally include non–ImageNet-based SD3.5 samples—present in AIGI-Holmes—since SD3.5 forms an isolated cluster and requires domain coverage in the curated set.}
\label{fig:mmd-threepanel}
\end{figure*}

\vspace{0.05in}
\noindent
{\bf ImageNet-domain hyperclusters.}
Figure~\ref{fig:mmd-threepanel} shows three hyperclusters among ImageNet-domain generators: (BigGAN \citep{brock2018large}, GLIDE \citep{nichol2021glide}), (SD1.4 \citep{rombach2022high}, Midjourney \citep{midjourney2024v61}), and an isolated cluster (SD3.5 \citep{esser2024scaling}). Although SD3.5 in AIGI-Holmes is not ImageNet-based, we generate auxiliary ImageNet-like SD3.5 samples to analyze its position and confirm its isolation. We include the original SD3.5 samples in the curated set due to their distinct distributional signature.

\vspace{0.05in}
\noindent
{\bf COCO-domain generators.}
The same hypercluster analysis is applied to COCO-domain generators, ensuring that each semantic domain contributes a compact yet distributionally diverse subset. This procedure removes redundant generators while preserving meaningful stylistic variation across domains.

\subsection{Integration of Supplemental Real Images}



\begin{table}
\centering
\small
\caption{
\textbf{Summary of generative models used in the AIGI-Holmes training split.}
We categorize models by their primary image domain (e.g., COCO2017, ImageNet, LSUN) 
and list all synthetic generators included in each domain group.
}
\label{tab:aigi-holmes-statistics}
\vspace{4pt}

\renewcommand{\arraystretch}{1}

\begin{tabular}{p{0.16\linewidth} 
                p{0.08\linewidth} 
                p{0.72\linewidth}}
\toprule
\textbf{Domain} & \textbf{\#Models} & \textbf{T2I Generation Models} \\
\midrule

\textbf{COCO2017} & 11 &
LCM-Lora-SD1.5 \citep{luo2023lcm},
LCM-Lora-SDXL \citep{luo2023lcm},
SD21-ControlNet-Canny \citep{zhang2023adding},         
SD-Turbo(SD2.1 Distilled),
StableDiffusion-v1.4 \citep{rombach2022high},
StableDiffusion-v1.5 \citep{rombach2022high},
StableDiffusion-v2.1 \citep{rombach2022high},
SDXL-Base-1.0 \citep{podell2023sdxl},
SDXL-Refiner-1.0,
SDXL-Inpainting-1.0,
SDXL-Turbo 
\\
\midrule

\textbf{ImageNet} & 8 &
BigGAN \citep{brock2018large},
VQDM \citep{gu2022vector}),
Glide \cite{nichol2021glide},
Midjourney,
StableDiffusion-v1.5,
StableDiffusion-v2.1,
ADM \citep{dhariwal2021diffusion},
Wukong \citep{wukong_modelzoo}
\\
\midrule

\textbf{LSUN} & 1 &
ProGAN \citep{karras2018progressive}
\\
\midrule

\textbf{Additional Domains} & 8 &
FLUX.1-dev \citep{blackforestlabs2024flux1dev},
StableDiffusion-v2.0,
StableDiffusion-v3.0-Medium \citep{esser2024scaling},
StableDiffusion-v3.5-Large \citep{esser2024scaling},
SDXL-Base,
SDXL-Turbo,
HumanRefiner\cite{fang2024humanrefiner},
Kandinsky \cite{razzhigaev2023kandinsky}
\\

\bottomrule
\end{tabular}
\end{table}
\begin{figure}[t!]
\centering
\begin{subfigure}{0.845\linewidth}
    \centering
    \includegraphics[width=\linewidth]{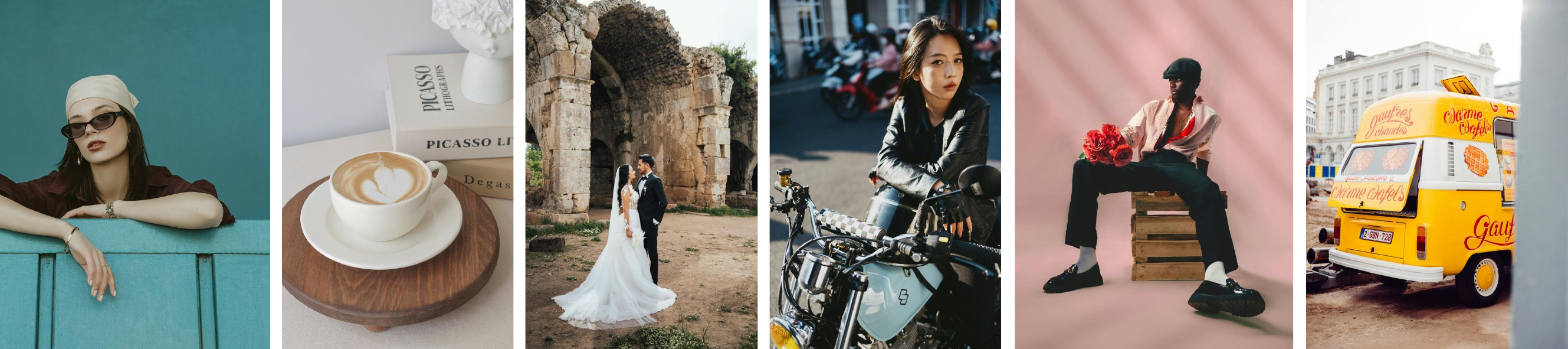}
    \caption{Pexels}
    \label{fig:suppreal-pexels}
\end{subfigure}
\hfill
\begin{subfigure}{0.85\linewidth}
    \centering
    \includegraphics[width=\linewidth]{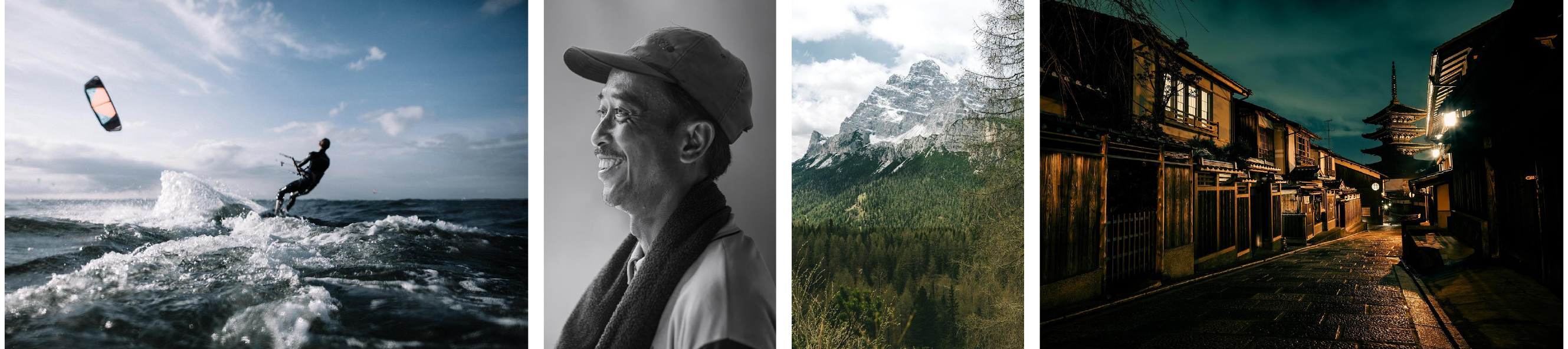}
    \caption{Pixabay}
    \label{fig:suppreal-pixabay}
\end{subfigure}
\hfill
\begin{subfigure}{0.85\linewidth}
    \centering
    \includegraphics[width=\linewidth]{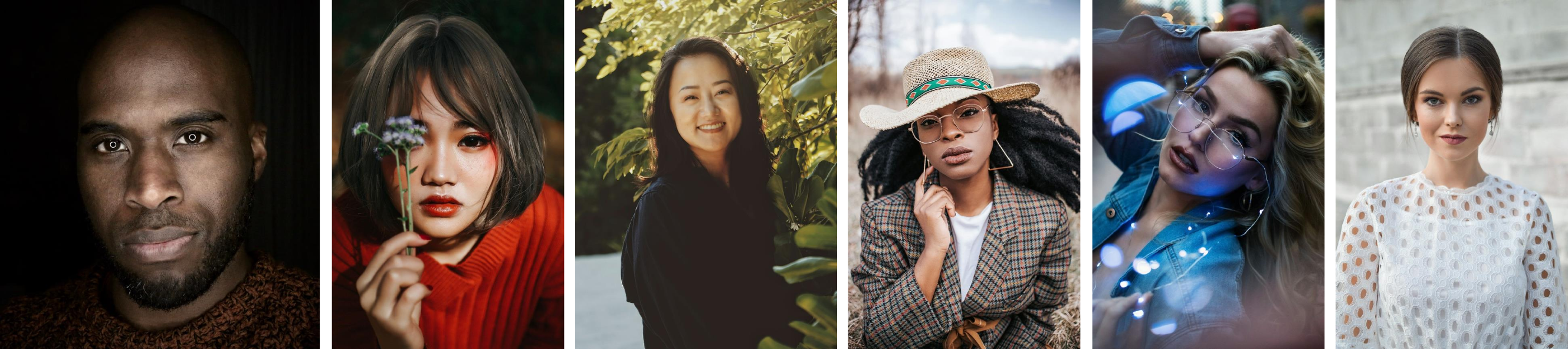}
    \caption{Unsplash (Portraits)}
    \label{fig:suppreal-unsplash}
\end{subfigure}

\vspace{-0.1in}
\caption{\textbf{Supplemental real-image examples.}  
Representative samples collected from Pexels, Pixabay, and Unsplash.  
These images provide high-resolution, high-quality object-centric, scene-centric, and portrait-centric real photographs that expand the real-image distribution used during training.}
\label{fig:suppreal-examples}
\end{figure}

To enrich the real-image distribution used for training, we incorporate an additional set of natural photographs that is distinct from the RealWorldBench evaluation images.  
While RealWorldBench focuses on modern high-quality smartphone images and commercial T2I outputs for benchmarking, the supplemental real images described here are collected purely for training diversity and are never used for evaluation.

We aggregate real photographs from four large-scale sources:
\vspace{0.03in}

\begin{itemize}[leftmargin=12pt]
    \item \textbf{Pexels}\footnote{\url{https://pexels.com}} and \textbf{Pixabay\footnote{\url{https://pixabay.com/}}}:  
    broad-coverage stock photography spanning indoor scenes, landscapes, products, food, wildlife, and everyday environments without any content restrictions.

    \item \textbf{Unsplash (Kaggle High-Quality Face Dataset)}:  
    high-resolution portrait images sourced from the public Unsplash face dataset\footnote{\url{https://www.kaggle.com/datasets/subrahmanya090/face-images-high-quality-scraped-from-unsplash}},  
    providing diverse identities, lighting conditions, and capture styles essential for improving robustness to face-centric real images.

    \item \textbf{Open Images V7\footnote{\url{https://storage.googleapis.com/openimages/web/index.html}}}:  
    large-scale, class-diverse natural images that further expand scene variety and improve the coverage of real-world object and background distributions.
\end{itemize}
\vspace{0.03in}

These sources contribute high-quality, high-resolution natural photographs across diverse real-world domains—including object-, scene-, and portrait-centric imagery—significantly expanding the real-image distribution beyond AIGI-Holmes.  
Incorporating these images during training allows the detector to better capture the real-image manifold and generalize more reliably to high-quality, high-resolution photographs encountered in real-world settings.  
Representative examples of these supplemental real images are shown in Figure~\ref{fig:suppreal-examples}.

\begin{figure}
    \centering
    \begin{subfigure}{0.46\linewidth}
        \centering
        \includegraphics[width=\linewidth]{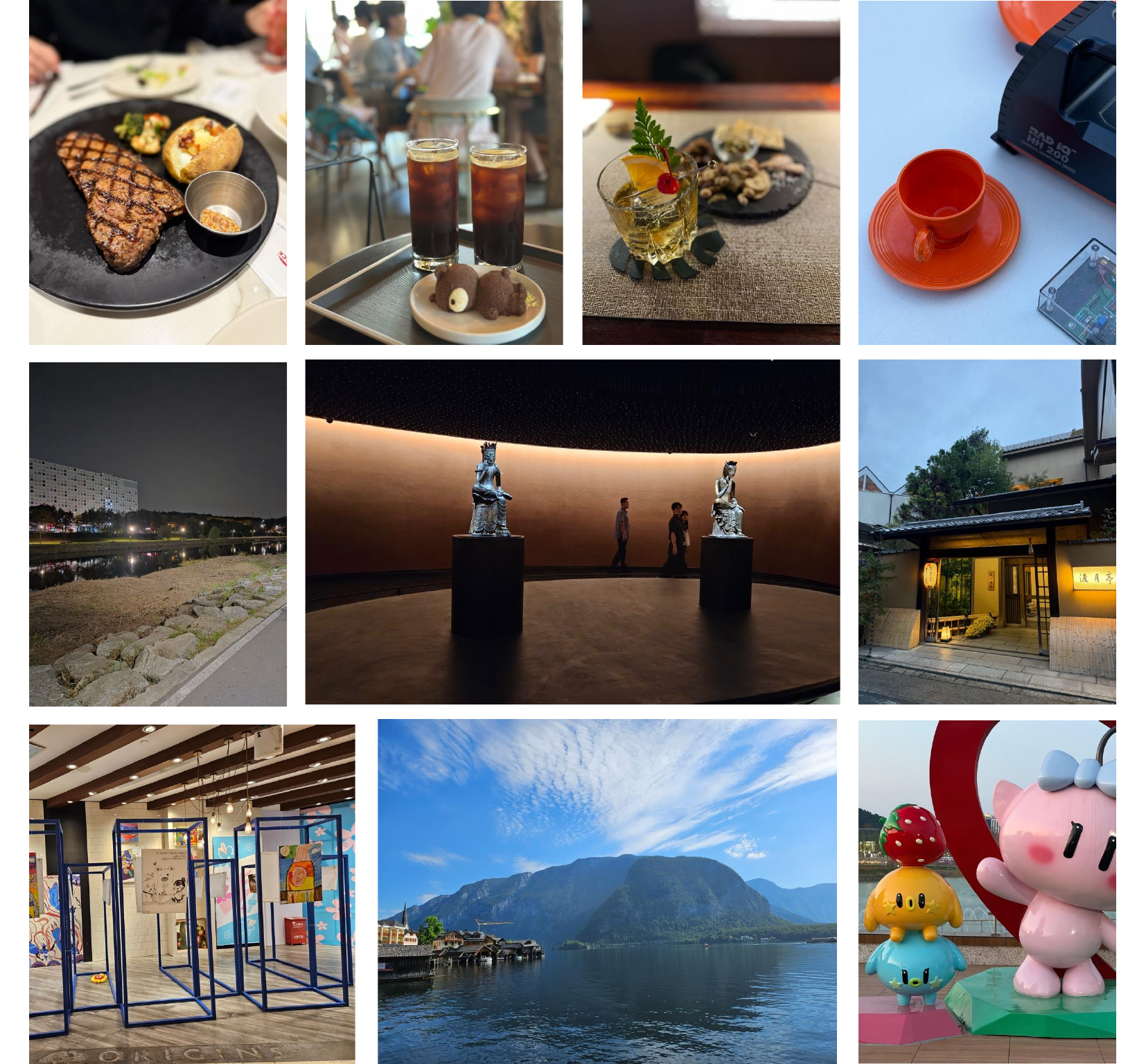}
        \caption{Smartphone-captured real images}
        \label{fig:realworldbench-smartphone}
    \end{subfigure}
    \hfill
    \begin{subfigure}{0.52\linewidth}
        \centering
        \includegraphics[width=\linewidth]{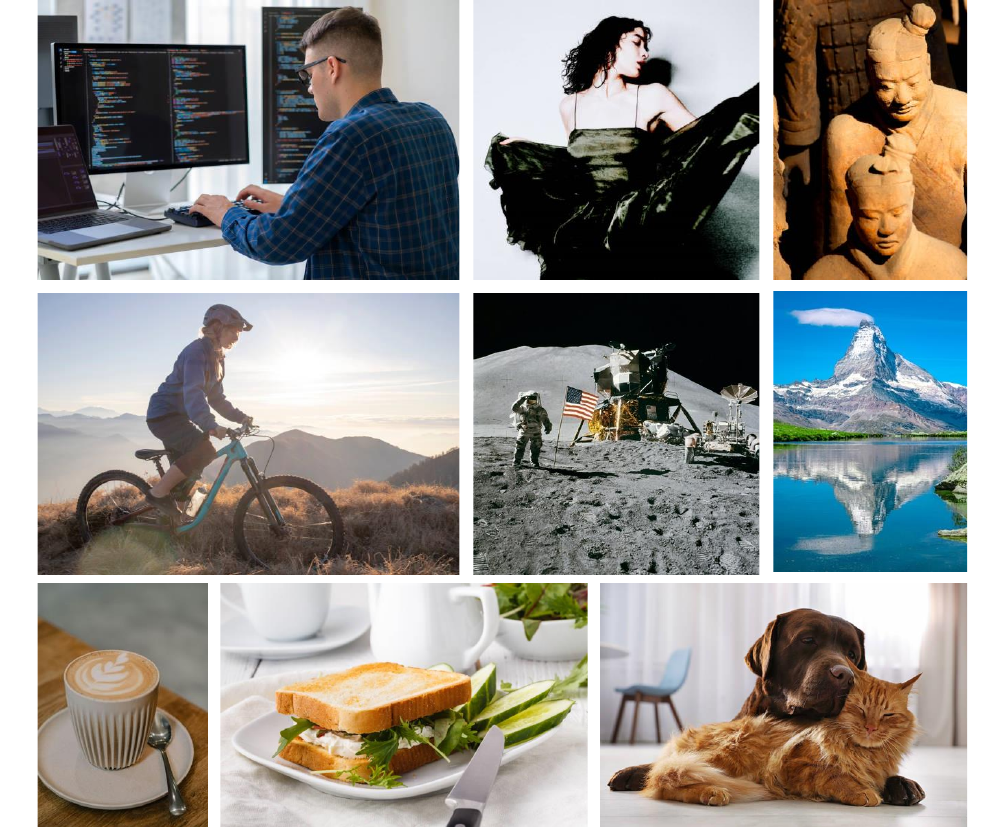}
        \caption{Web-sourced real images}
        \label{fig:realworldbench-web}
    \end{subfigure}

    \caption{\textbf{RealWorldBench real-image examples.} 
    Representative high-resolution real images from (a) modern smartphone cameras and (b) curated high-quality web sources. 
    These images illustrate the distribution shift relative to commonly used datasets such as ImageNet or LAION.}
    \label{fig:realworldbench-real-examples}
\end{figure}
\begin{table}[t!]
\centering
\small
\caption{\textbf{Composition of the Final Curated Dataset.}
We construct a compact yet diverse 10K training set, consisting of 5K real and 5K fake images collected from heterogeneous sources.}
\vspace{4pt}

\begin{tabular}{cc}
\toprule
\textbf{Real Image Sources} & \textbf{Fake Image Sources} \\
\midrule
LSUN & SD1.4 (ImageNet domain) \\
ImageNet & BigGAN (ImageNet domain) \\
COCO 2017 & GLIDE (ImageNet domain) \\
OpenImages V7 & Midjourney (ImageNet domain) \\
Unsplash People & SDXL-LoRA (COCO domain) \\
Pexels & SD2.1 (COCO domain) \\
Pixabay & SD1.5 (COCO domain) \\
& SD3.5 (Additional; non–ImageNet) \\
\bottomrule
\end{tabular}
\label{tab:curated_dataset}
\end{table}

\subsection{Final Curated Dataset}
Our curation procedure yields a compact but distributionally rich training set composed of
{approximately 10K carefully curated images}, as summarized in Table~\ref{tab:curated_dataset}.  
The dataset is constructed to preserve semantic diversity, generator-level coverage, and domain heterogeneity across both real and synthetic samples.

This balanced composition ensures that the curated training set captures  
(1) high intra-real diversity across real-world domains,  
(2) hypercluster-level variation across synthetic generators, and  
(3) mitigating the common issue where detectors overfit to specific real or fake datasets and fail on new ones.


\section{RealWorldBench: Test Benchmark Details}
\label{sec:appendix_realworldbench}

RealWorldBench is a high-fidelity, test-only benchmark designed to evaluate the generalization ability of AI-generated image detectors under realistic conditions. It consists of two components: (1) real images captured with modern smartphone cameras and collected from high-quality web sources, and (2) synthetic images produced by a wide range of recent commercial and open-source T2I models.  
Representative real-image examples are shown in Figure~\ref{fig:realworldbench-real-examples}.

\subsection{Real Image Composition}

\subsubsection{Smartphone-Captured Real Images}
A substantial portion of the real images in RealWorldBench consists of over 11K photographs captured using modern smartphone devices, spanning both iPhone and Samsung Galaxy product lines.  

\begin{table}
\centering
\caption{Resolution statistics for smartphone-captured real images in RealWorldBench. 
Each device is summarized by its most frequent (representative) resolution.}
\vspace{-0.03in}
\label{tab:realworldbench_resolution_stats}
\resizebox{0.9\linewidth}{!}{
\begin{tabular}{lcc|lcc}
\toprule
\textbf{Smartphone} & Representative Resolution & \# Images  & \textbf{Smartphone} & Representative Resolution & \# Images  \\
\midrule
galaxy-a53        & 4624$\times$3468 & 40  & iphone-12-pro     & 3024$\times$4032 &98  \\
galaxy-note-9     & 4032$\times$3024 & 30  & iphone-13         & 4032$\times$3024 &19  \\
galaxy-s20+       & 3024$\times$4032 & 23  & iphone-13-pro     & 4032$\times$3024 &30  \\
galaxy-s21        & 4032$\times$3024 & 20  & iphone-14-pro     & 4032$\times$3024 &30  \\
galaxy-s22        & 4000$\times$3000 & 15  & iphone-15         & 5712$\times$4284 &131  \\
galaxy-s23        & 4000$\times$3000 & 233  & iphone-15-plus    & 4284$\times$5712 &158  \\
galaxy-s23-plus   & 4000$\times$3000 & 30  & iphone-16-pro     & 3024$\times$4032 &140  \\
galaxy-s25        & 4000$\times$3000 & 14  & iphone-17         & 5712$\times$4284 &10  \\
galaxy-s25-plus   & 8160$\times$6120 & 15  & iphone-17-pro     & 4032$\times$3024 &17  \\
& & & iphone-se-3       & 4032$\times$3024 &10  \\
& & & iphone-xr         & 3024$\times$4032 &30  \\
\bottomrule
\end{tabular}
}
\end{table}

\begin{table}[t!]
\centering
\caption{Number of synthetic images per T2I generator included in RealWorldBench.}
\label{tab:realworldbench_fake_models}
\small
\begin{tabular}{cc|cc|cc}
\toprule
\textbf{Model} & \# Images & \textbf{Model} & \# Images &
\textbf{Model} & \# Images \\
\midrule
sd14 & 1128 & aurora-20-1-25 & 1132 & imagen-3 & 1075 \\
sd21 & 1128 & flux\_dev & 1128 & imagen4 & 1128 \\
sd3\_medium & 1128 & flux\_krea & 1128 & imagen-4-ultra-24-7-25 & 1029 \\
sd35\_large & 1128 & flux-1-pro & 1124 & lumina-17-2-25 & 589 \\
kolors & 1128 & flux-1.1-pro & 1123 & midjourney-5.2 & 1091 \\
sd3\_medium & 1128 & frames-23-1-25 & 1002 & recraft-v2 & 848 \\
sdxl\_base & 1128 & halfmoon-4-4-25 & 761 & recraft-v3-24-7-25 & 1132 \\
sdxl\_turbo & 1128 & hidream-I1-full & 857 & seedream-3-24-7-25 & 1124 \\
dalle-3 & 1082 & ideogram & 849 & 4o-26-3-25 & 998 \\
janus-7b & 1132 & nano\_banana & 1124 & ideogram & 849 \\
\bottomrule
\end{tabular}
\end{table}

Table~\ref{tab:realworldbench_resolution_stats} summarizes the number of images collected per device, along with their representative resolutions.  
The resulting distribution covers a broad range of image sizes (typically 3,000--4,500px on the long side), natural sensor noise patterns, optical blur, and diverse lighting conditions frequently encountered in everyday smartphone photography.

\subsubsection{Web-Based High-Quality Real Images}
To supplement smartphone photographs and further enhance domain diversity, RealWorldBench includes additional high-quality real images sourced from \textit{Pexels} and \textit{iStockPhoto}.  
We curate balanced content across various scene categories, including portraits, landscapes, food, indoor environments, outdoor settings, products, and wildlife.  
A subset of web images is additionally collected using prompts from the Rapidata Text-to-Image Human Preference dataset \citep{rapidata_recraft_v3_2025}.  
Low-resolution, heavily compressed, or watermarked images are systematically filtered out to maintain benchmark fidelity.

\subsection{AI-Generated Image Composition}

\subsubsection{Text-to-Image Generators}
RealWorldBench includes synthetic images generated from a wide spectrum of modern T2I models.  
Table~\ref{tab:realworldbench_fake_models} reports the number of images generated per model.

The generators span diverse model families, including diffusion transformers, autoregressive image models, large-scale proprietary systems, and recent open-source baselines.  
This variety ensures comprehensive coverage of generator-induced domain shifts and stylistic variations.

\subsubsection{Prompt-Aligned Sampling via Rapidata Human Preference Dataset}
A core component of RealWorldBench is the use of the \textit{Text-to-Image Human Preference Dataset} released by Rapidata.
For each of the 282 prompts, multiple generators produce 3–4 semantically aligned images, enabling controlled analysis of generator-specific characteristics while minimizing semantic drift.
This design allows fair comparison across models under identical prompts and supports evaluating detectors in diverse yet semantically matched conditions. Example prompts are shown below.
\vspace{0.05in}
\begin{itemize}[leftmargin=12pt]
    \item Hyperrealism, man and woman, together, made of clay.
    \item Two hot dogs sit on a white paper plate near a soda cup on a green picnic table, with a bike and a silver car parked nearby.
    \item A blue and white cat next to a blanket and a shelf with a grey bottle.
\end{itemize}
\vspace{0.05in}
These examples illustrate the diversity of semantics and scene structures covered in the Rapidata dataset, which enables consistent cross-generator comparisons within RealWorldBench.

\begin{table*}[ht!]
\centering
\caption{
\textbf{Generalization performance across diverse synthetic generators and real-image benchmarks.}
All detectors are trained on images generated by SD-v1.4 (144K) and ProGAN (144K).  
We compare four training sets of increasing compactness—AIGIBench (288K), AIGI-Holmes (45K), Universal (50K), and our Optimized subset (10K)—and evaluate their robustness across AIGIBenchmark including two real-image domains (SocialRF, CommunityAI).  
Results demonstrate that our \textbf{10K optimized subset} maintains competitive or superior generalization to unseen generators despite being significantly smaller.
}
\label{tab:aigibench_curation}
\vspace{6pt}

\small
\renewcommand\arraystretch{1.1}
\setlength{\tabcolsep}{4pt}

\resizebox{\linewidth}{!}{
\begin{tabular}{lcc|cc|cc|cc|cc|cc|cc|cc|cc}
\toprule
\bf Test →
& \multicolumn{2}{c}{ProGAN}
& \multicolumn{2}{c}{R3GAN}
& \multicolumn{2}{c}{StyleGAN3}
& \multicolumn{2}{c}{StyleGAN-XL}
& \multicolumn{2}{c}{StyleSwin}
& \multicolumn{2}{c}{WFIR}
& \multicolumn{2}{c}{BlendFace}
& \multicolumn{2}{c}{E4S}
& \multicolumn{2}{c}{FaceSwap}
\\
\cmidrule{2-19}
\bf Train ↓
& Acc&AP & Acc&AP & Acc&AP & Acc&AP & Acc&AP & Acc&AP & Acc&AP & Acc&AP & Acc&AP \\
\midrule
AIGIBench (288K)   &100&100&98.2&100&97.9&99.9&98.3&100&98.1&99.9&95.3&100&62.3&50.1&82.9&67.7&83.0&57.0\\
AIGI-Holmes (45K)  &99.8&100&96.6&100&97.9&99.9&95.9&100&96.5&99.6&96.8&100&49.7&52.6&67.9&82.0&59.0&71.9\\
Universal (50K)    &99.7&100&98.3&100&98.7&100&98.4&100&98.4&99.9&99.5&100&52.4&67.6&75.7&93.5&60.1&78.9\\
Optimized (10K)    &99.3&100&98.9&100&99.0&100&98.7&100&96.7&100&99.8&100&52.2&76.2&70.0&93.9&57.4&80.1\\
\bottomrule
\end{tabular}}

\resizebox{\linewidth}{!}{
\begin{tabular}{lcc|cc|cc|cc|cc|cc|cc|cc|cc}
\toprule
\bf Test →
& \multicolumn{2}{c}{InSwap}
& \multicolumn{2}{c}{SimSwap}
& \multicolumn{2}{c}{Flux1-dev}
& \multicolumn{2}{c}{Midjourney-V6}
& \multicolumn{2}{c}{GLIDE}
& \multicolumn{2}{c}{DALLE-3}
& \multicolumn{2}{c}{Imagen3}
& \multicolumn{2}{c}{SD3}
& \multicolumn{2}{c}{SDXL}
\\
\cmidrule{2-19}
\bf Train ↓
&Acc&AP&Acc&AP&Acc&AP&Acc&AP&Acc&AP&Acc&AP&Acc&AP&Acc&AP&Acc&AP\\
\midrule
AIGIBench (288K)&84.0&64.7&83.8&63.6&98.2&99.9&93.9&99.1&97.4&99.7&91.2&98.7&93.8&98.9&98.4&100&98.8&100\\
AIGI-Holmes (45K)&54.4&61.8&57.0&70.6&97.3&100&95.5&99.9&97.1&100&96.8&99.9&95.4&99.3&96.4&100&97.4&100\\
Universal (50K)&55.7&72.2&60.0&82.5&98.4&99.9&97.9&99.8&98.3&100&98.9&99.8&84.5&95.3&98.5&100&99.1&100\\
Optimized (10K)&53.5&68.8&56.4&81.1&99.1&100&98.8&100&99.2&100&98.4&99.8&84.9&96.6&98.9&100&99.3&100\\
\bottomrule
\end{tabular}}

\resizebox{\linewidth}{!}{
\begin{tabular}{lcc|cc|cc|cc|cc|cc|cc|cc}
\toprule
\bf Test →
& \multicolumn{2}{c}{BLIP}
& \multicolumn{2}{c}{Infinite-ID}
& \multicolumn{2}{c}{InstantID}
& \multicolumn{2}{c}{IP-Adapter}
& \multicolumn{2}{c}{PhotoMaker}
& \multicolumn{2}{c}{SocialRF}
& \multicolumn{2}{c}{CommunityAI}
& \multicolumn{2}{c}{\cellcolor{lightgray!20}\textbf{Mean}}
\\
\cmidrule{2-17}
\bf Train ↓
&Acc&AP&Acc&AP&Acc&AP&Acc&AP&Acc&AP&Acc&AP&Acc&AP
&\cellcolor{lightgray!20}\textbf{Acc}&\cellcolor{lightgray!20}\textbf{AP}\\
\midrule
AIGIBench (288K)&99.8&100&98.3&100&98.3&100&98.4&100&96.7&100&83.3&92.5&73.9&99.3
&\cellcolor{lightgray!20}88.5&\cellcolor{lightgray!20}95.3\\
AIGI-Holmes (45K)&97.6&100&96.6&100&96.6&100&96.9&100&96.4&99.8&70.7&95.3&82.8&99.5
&\cellcolor{lightgray!20}87.4&\cellcolor{lightgray!20}93.3\\
Universal (50K)&98.4&100&98.9&100&98.5&100&98.4&99.9&98.6&99.7&96.3&99.1&94.9&98.7
&\cellcolor{lightgray!20}\textbf{90.3}&\cellcolor{lightgray!20}95.5\\
Optimized (10K)&99.1&100&99.6&100&99.2&100&98.3&99.8&94.6&99.1&91.6&98.5&92.6&99.0&\cellcolor{lightgray!25}89.4&\cellcolor{lightgray!25}\textbf{95.7}\\
\bottomrule
\end{tabular}}
\end{table*}

\section{Ablation Studies}\label{appendix_c}
\subsection{Effectiveness of Representation-Aware Dataset Optimization}

Table~\ref{tab:aigibench_curation} summarizes how different training sets influence generalization across diverse generators. Models trained on the full AIGIBench dataset (288K images) perform well on most fully synthetic generators (e.g., ProGAN, R3GAN, StyleGAN-family, diffusion models), but their accuracy drops significantly on identity-preserving editing pipelines such as BlendFace, E4S, FaceSwap, SimSwap, and InSwap. These pipelines alter only localized facial regions while retaining most of the real background, producing hybrid images with subtle and spatially confined artifacts that are difficult to detect. This indicates that large-scale datasets alone do not provide robustness to partially generated content.

The proposed {Optimized-10K} subset, selected via representation-aware filtering, achieves comparable or better generalization while being nearly $30\times$ smaller. It maintains strong performance not only on GAN and diffusion models but also on modern commercial generators (e.g., Midjourney-V6, Imagen3, FLUX-1-dev, SD3, SDXL) and transfers well to real-world benchmarks such as SocialRF and CommunityAI.  

Overall, these results show that a compact, carefully curated training set can be as effective as large-scale datasets, highlighting the importance of sample quality and semantic diversity over sheer data volume.


\subsection{Ablation on Vision Encoder Architecture}

Table~\ref{tab:enc_ablation} compares PE-Linear with different frozen vision encoders. All models are trained on a 40K subset of the AIGIBench training split and evaluated on the AIGIBench test set (excluding face-swapping datasets). Multimodal encoders (PE-Core and SigLIP2) substantially outperform self-supervised encoders (DINOv2 and DINOv3), indicating that multimodal pretraining provides more discriminative features for real/fake classification. Among all backbones, PE-Core achieves the highest accuracy (95.4\%) and AP (99.4), making it the most effective frozen encoder in our setting.

\subsection{Encoder Choice for Dataset Curation vs. Downstream Training}
\begin{figure}[t!]
\centering
\vspace{-0.05in}

\centering\small
\captionof{table}{
\textbf{Ablation on frozen vision encoder architecture.}
We compare PE-Linear across various pretrained vision encoders to examine how architectural properties 
(multimodal PE and SigLIP vs. self-supervised DINO) affect real/fake discrimination.
All models are trained on the AIGIBench training split using a 40K subsampled dataset, and tested with AIGIBench except face-swapping datasets.
}
\vspace{-0.05in}

\resizebox{0.5\linewidth}{!}{
\begin{tabular}{lcccc}
\toprule
Frozen Vision Encoder & Real Acc. & Fake Acc. & Acc. & A.P. \\
\midrule
PE-Core-G14-448   & \textbf{95.1} & \textbf{95.7} & \textbf{95.4} & \textbf{99.4} \\
SigLIP2-G16-384   & 85.5 & 89.3 & 87.4 & 95.5 \\
DINOv3 ViT-L/16   & 79.9 & 87.7 & 83.8 & 92.7 \\
DINOv2 Giant      & 76.0 & 69.5 & 72.8 & 78.1 \\
\bottomrule
\end{tabular}
\label{tab:enc_ablation}
}
\end{figure}
\begin{table}[t!]
\centering
\vspace{-0.05in}

\centering\small
\captionof{table}{
\textbf{Ablation on encoder choice for dataset optimization versus downstream training.}
We study whether the vision encoder used for (1) selecting informative training samples and 
(2) training the final classifier should be identical or different. 
The optimized training subset is constructed from the AIGI-Holmes training set combined with web-crawled real images (Pixabay and Pexels).
}
\vspace{-0.05in}

\resizebox{0.5\linewidth}{!}{
\begin{tabular}{l|cccc}
  
\toprule
  \textbf{Data Curating Encoder}&\multicolumn{4}{c}{PE-Core-G14-448}\\
 \midrule
 \textbf{Backbone Encoder} & Real Acc. & Fake Acc. & Acc. & A.P. \\
\midrule
 PE-Core-G14-448 & \textbf{98.6}& \textbf{80.1}& \textbf{89.4}& \textbf{95.7}\\
 SigLIP2-G16-384 & 93.1& 69.4& 81.3& 90.4\\
 DINOv3 ViT-L/16 & 94.4& 63.6& 79.0& 90.1\\
 DINOv2 Giant    & 93.4& 34.8& 64.2&77.2\\

 \bottomrule
\end{tabular}
}
\vspace{0.01in}

\resizebox{0.5\linewidth}{!}{
\begin{tabular}{l|cccc}

\toprule
  \textbf{Data Curating Encoder}&\multicolumn{4}{c}{SigLIP2-G16-384}\\
 \midrule
 \textbf{Backbone Encoder} & Real Acc. & Fake Acc. & Acc. & A.P. \\
\midrule
 PE-Core-G14-448 & \textbf{98.2}& \textbf{78.8}& \textbf{88.5}& \textbf{95.1}\\
 SigLIP2-G16-384 & 94.6& 69.5& 82.1& 90.7\\
 DINOv3 ViT-L/16 & 94.7& 68.1& 81.4& 91.7\\
 DINOv2 Giant    & 92.9& 33.8& 63.4&76.9\\

 \bottomrule
\end{tabular}
}

\vspace{0.01in}

\resizebox{0.5\linewidth}{!}{
\begin{tabular}{l|cccc}

\toprule
  \textbf{Data Curating Encoder}&\multicolumn{4}{c}{Random Selecting}\\
 \midrule
 \textbf{Backbone Encoder} & Real Acc. & Fake Acc. & Acc. & A.P. \\
\midrule
 PE-Core-G14-448 & \textbf{99.5}& \textbf{73.2}& \textbf{86.3}& \textbf{94.1}\\
 SigLIP2-G16-384 & 96.4& 62.6& 79.5& 90.2\\
 DINOv3 ViT-L/16 & 94.3& 65.9& 80.1& 90.7\\
 DINOv2 Giant    & 93.8& 32.7& 63.3&74.7\\

 \bottomrule
\end{tabular}
}
\label{tab:ablation-encoder-choices}
\end{table}

To quantify how encoder choice affects both dataset construction and downstream classification, we evaluate all combinations of four encoders—PE-Core-G14-448, SigLIP2-G16-384, DINOv3 ViT-L/16, and DINOv2-Giant—used respectively for (1) dataset curation and (2) training the final linear classifier. All models are trained on the curated 10K training subset and evaluated on the full AIGIBench test set. The results in Table~\ref{tab:ablation-encoder-choices} reveal several consistent trends. 

First, PE-Core consistently achieves the strongest downstream performance across curation strategies, suggesting that multimodal representations yield more discriminative features for real/fake classification than self-supervised representations.

Second, performance is generally highest when the same encoder is used for both dataset curation and downstream training. For example, PE-Core achieves its best result when used for both stages (Acc = 89.4, AP = 95.7), while SigLIP2 similarly benefits from encoder alignment. This suggests that selecting training samples in the same representation space as the classification model reduces distribution mismatch.

Finally, representation-aware curation consistently outperforms random selection across all encoder choices. The effect is particularly noticeable for multimodal encoders, indicating that both encoder quality and encoder alignment play important roles in constructing effective training subsets.

\begin{figure}
\centering\small
\vspace{-0.3in}
\includegraphics[width=0.86\linewidth]{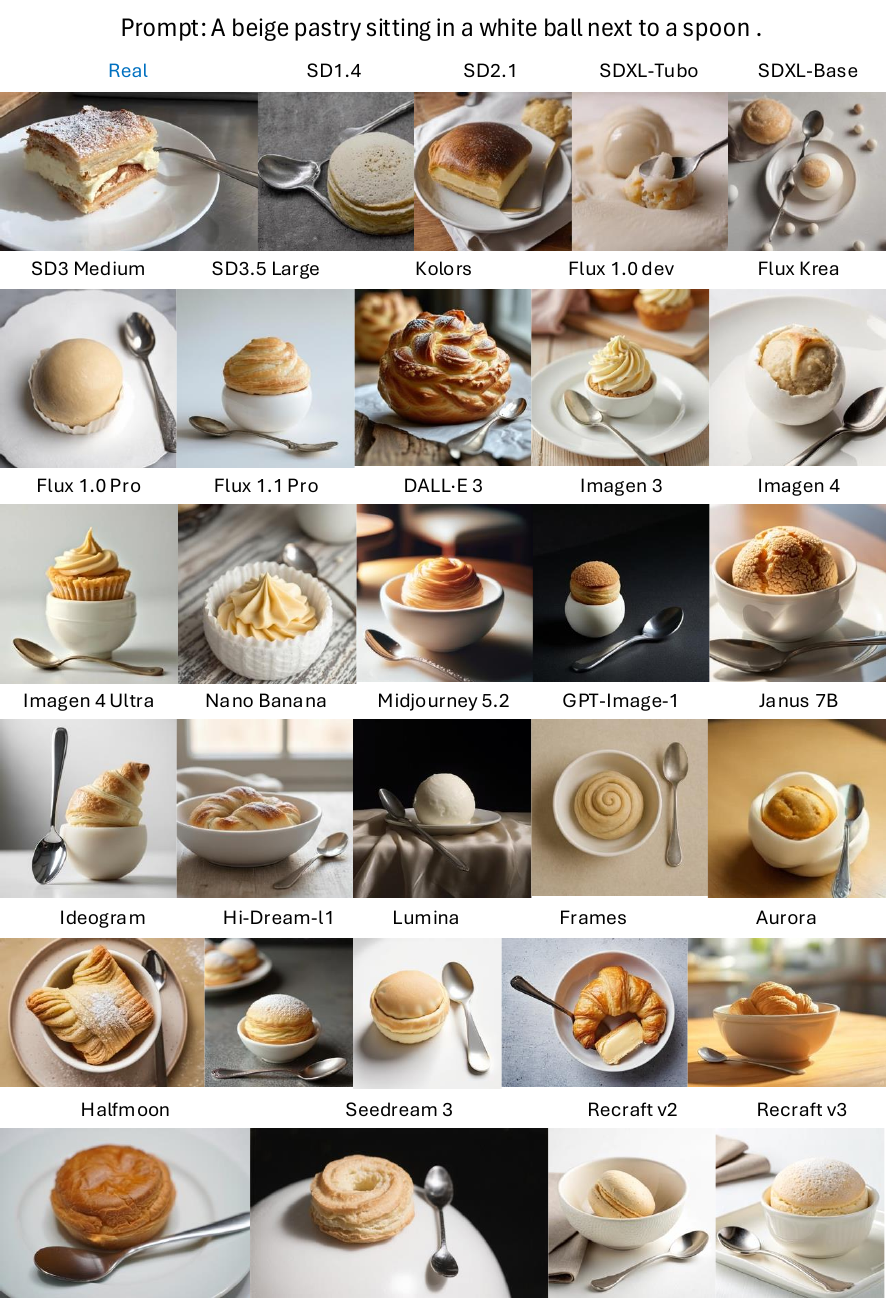}
\vspace{-0.03in}
\caption{\textbf{Example images from our newly constructed RealWorldBench.}
To minimize semantic variation across models, all synthetic images are generated using the \textit{same prompt} 
(“A beige pastry sitting in a white bowl next to a spoon”). 
The figure includes real photographs and outputs from 28 commercial and open-source text-to-image models, illustrating the wide stylistic and textural differences that persist even under identical semantic conditions.}
\label{fig:realworld_big_examples}
\end{figure}

\begin{figure}
\centering\small
\vspace{-0.3in}
\includegraphics[width=0.86\linewidth]{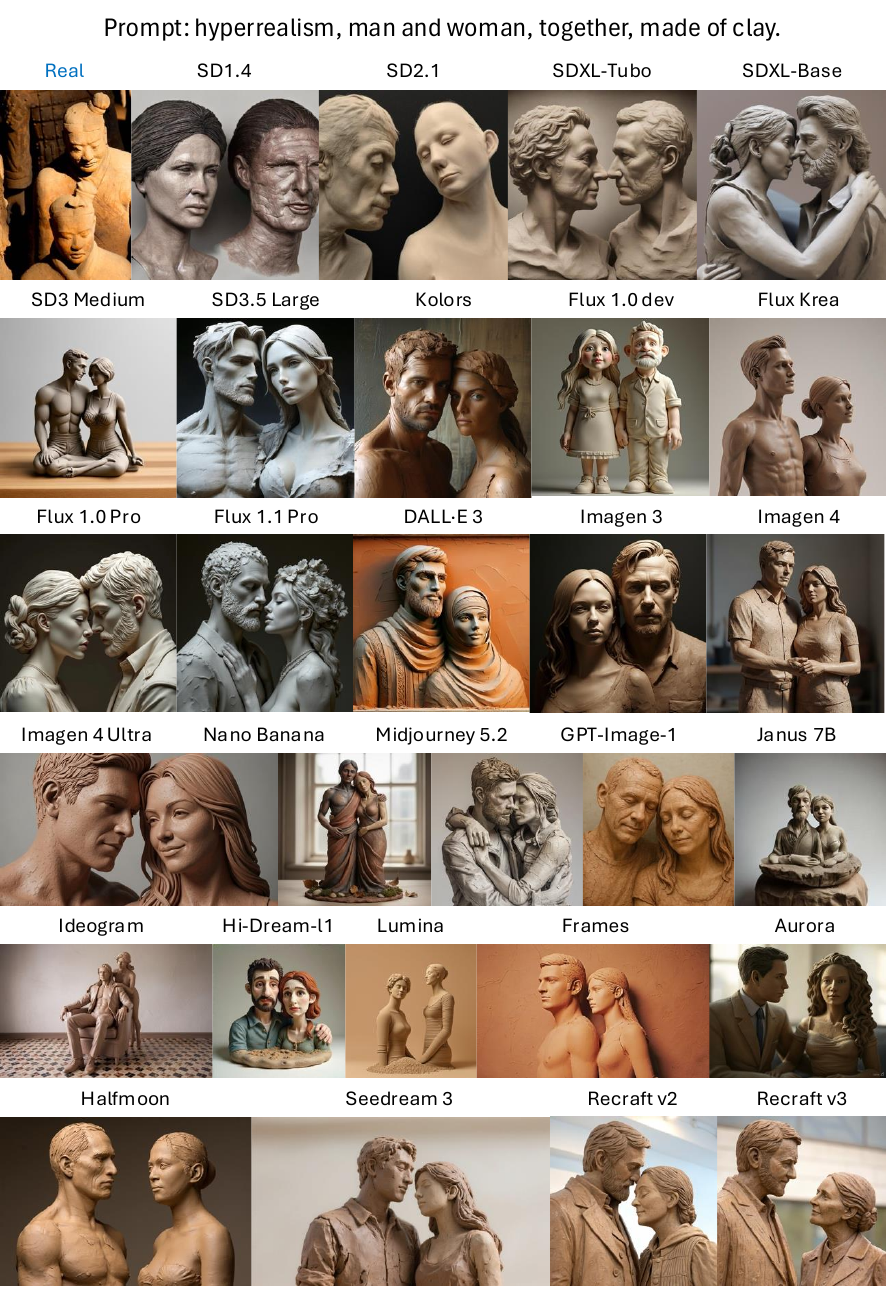}
\vspace{-0.03in}
\caption{\textbf{Example images from our newly constructed RealWorldBench.}
To minimize semantic variation across models, all synthetic images are generated using the \textit{same prompt} 
(“hyperrealism, man and woman, together, made of clay.”). 
The figure includes real photographs and outputs from 28 commercial and open-source text-to-image models, illustrating the wide stylistic and textural differences that persist even under identical semantic conditions.}
\label{fig:realworld_big_examples2}
\end{figure}



\end{document}